\documentclass{bmvc2k}

\usepackage{multirow}
\usepackage[normalem]{ulem}

\usepackage{mathtools}
\usepackage{amssymb}
\usepackage{amsmath}

\usepackage{makecell}

\newcommand{\argmax}{\mathop{\mathit{argmax}}\limits}

\usepackage{multicol}

\title{CEREALS -- Cost-Effective REgion-based Active Learning for Semantic Segmentation}
\addauthor{Radek Mackowiak}{radek.mackowiak@de.bosch.com}{1}
\addauthor{Philip Lenz}{philip.lenz@de.bosch.com}{1}
\addauthor{Omair Ghori}{omair.ghori@de.bosch.com}{1}
\addauthor{Ferran Diego}{ferran.diego@de.bosch.com}{1}
\addauthor{Oliver Lange}{oliver.lange@de.bosch.com}{1}
\addauthor{Carsten Rother}{carsten.rother@iwr.uni-heidelberg.de}{2}

\addinstitution{
 Robert Bosch GmbH \\ 
 Corporate Research - Computer Vision \\
 Robert-Bosch-Stra{\ss}e 200\\
 31139 Hildesheim, DE
}
\addinstitution{
 Heidelberg Collaboratory for Image Processing (HCI) \\
 Berliner Stra{\ss}e 43,\\
 69120 Heidelberg, DE
}

\runninghead{MACKOWIAK ET AL.}{CEREALS}

\def\etal{\emph{et al}\bmvaOneDot}

\begin{document}

\maketitle

\begin{abstract}
State of the art methods for semantic image segmentation are trained in a supervised fashion using a large corpus of fully labeled training images. However, gathering such a corpus is expensive, due to human annotation effort, in contrast to gathering unlabeled data. We propose an active learning-based strategy, called CEREALS, in which a human only has to hand-label a few, automatically selected, regions within an unlabeled image corpus. This minimizes human annotation effort while maximizing the performance of a semantic image segmentation method. The automatic selection procedure is achieved by: a) using a suitable information measure combined with an estimate about human annotation effort, which is inferred from a learned cost model, and b) exploiting the spatial coherency of an image. The performance of CEREALS is demonstrated on Cityscapes, where we are able to reduce the annotation effort to 17\%, while keeping 95\% of the mean Intersection over Union (mIoU) of a model that was trained with the fully annotated training set of Cityscapes.
\end{abstract}

\section{Introduction}

Deep convolutional neural networks (CNNs) have become the de-facto standard method for solving a large variety of heterogeneous image understanding problems. In the domain of visual scene understanding, semantic segmentation plays an important role due to enabling machines a pixel-wise semantic understanding of their environment. It is therefore a key enabler for applications like autonomous driving or robotic vision. However, one shortcoming of current CNN training algorithms is that they require a large amount of diverse and labeled training data to achieve satisfying results. Furthermore, their performance seems to scale linearly with an exponential increase of training data~\cite{DBLP:conf/iccv/SunSSG17}.

\begin{figure}[!ht]
\centering
\begin{subfigure}[Ground Truth \label{fig:teaser:gt}]{\includegraphics[width=0.24\textwidth]{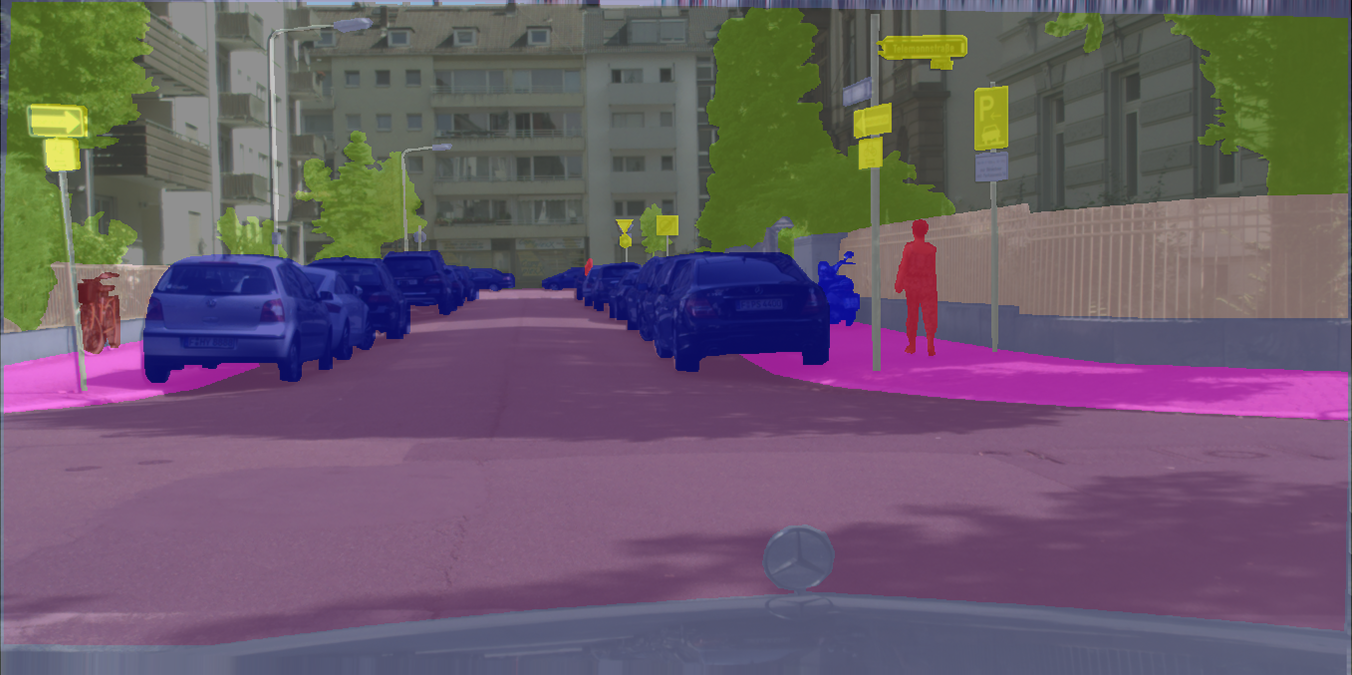}} \end{subfigure}
\begin{subfigure}[Full training set (100\% labeling effort) \label{fig:teaser:100_percent}]{\includegraphics[width=0.24\textwidth]{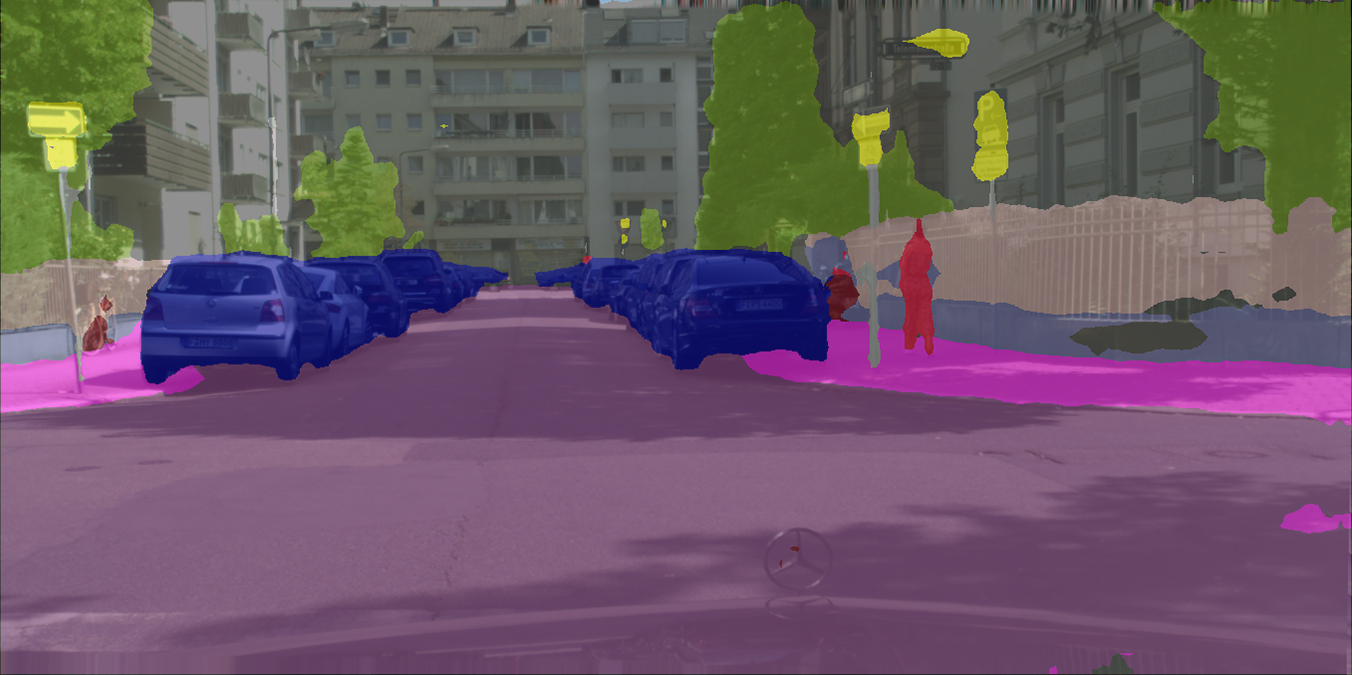}}  \end{subfigure}
\begin{subfigure}[$17\%$ of the effort by random data selection \label{fig:teaser:20_percent_random}]{\includegraphics[width=0.24\textwidth]{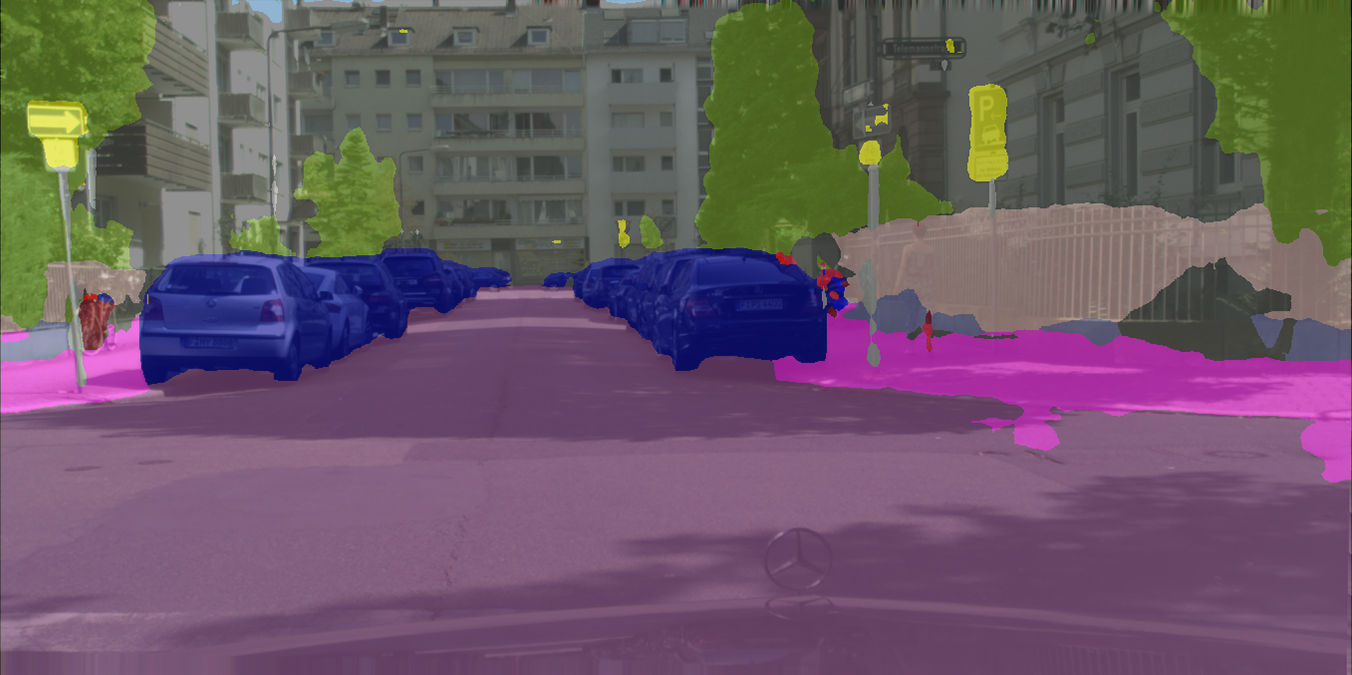}} \end{subfigure}
\begin{subfigure}[$17\%$ of the effort using our approach. \label{fig:teaser:20_percent_method}]{\includegraphics[width=0.24\textwidth]{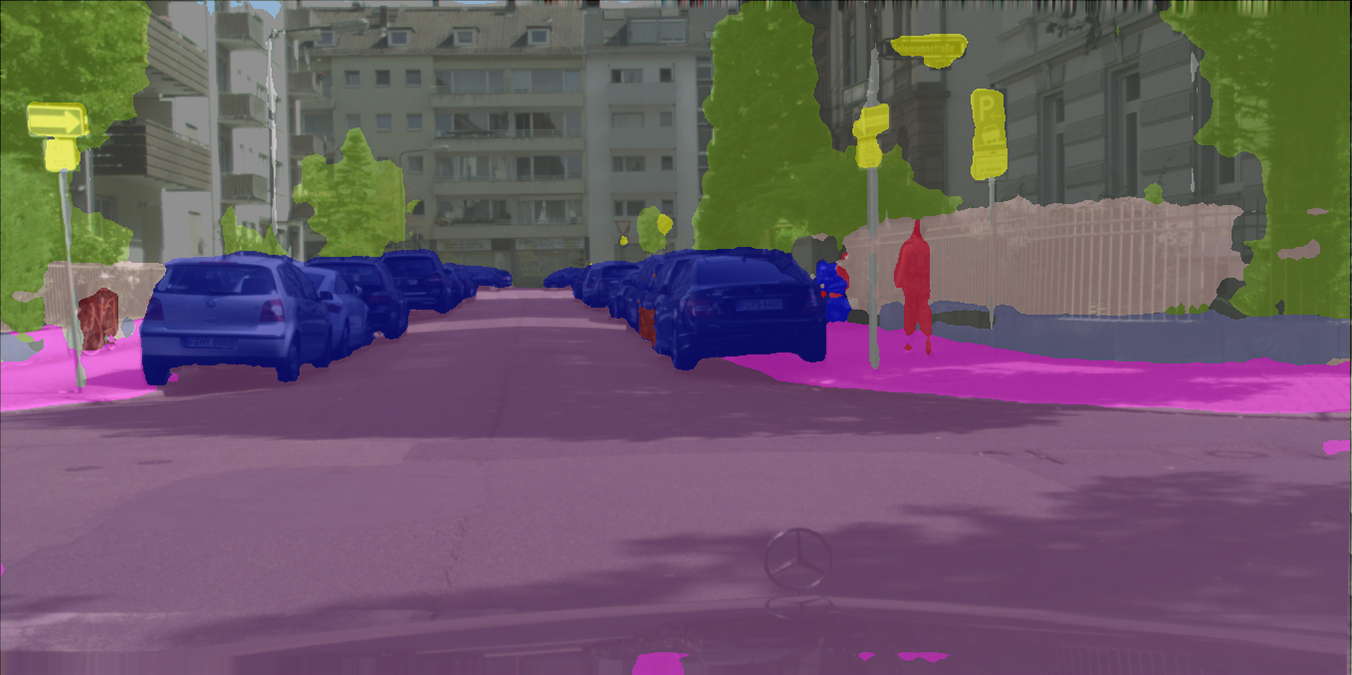}} \end{subfigure}
\caption{Qualitative segmentation results using \textit{CEREALS} for data annotation. Our approach reduces labeling effort significantly. We achieve $95\%$ of the performance with only $17\%$ of the labeling effort measured by the number of clicks as compared to annotating the full training set of Cityscapes \cite{DBLP:conf/cvpr/CordtsORREBFRS16}.}
\label{fig:teaser}
\vspace{-0.25cm}
\end{figure}

While acquiring large amounts of unlabeled data is usually easy, the effort required to manually annotate this data is a costly process due to the requirement of human annotators~\cite{DBLP:journals/ftcgv/KovashkaRFG16, DBLP:conf/cvpr/CordtsORREBFRS16}. Hence, the major bottleneck for rapidly applying CNN models into new domains is the acquisition of large-scale labeled training sets. Active Learning (AL)~\cite{DBLP:conf/nips/CohnGJ94} is an established approach to mitigate the problems associated with data labeling. In essence AL aims to query the data only for annotation, which is more likely to lead to more accurate models when used for training than other data~\cite{DBLP:journals/jmlr/TongK01, DBLP:conf/nips/TongK00,DBLP:conf/icml/GalIG17}. Consequently, this mitigates the time and monetary cost associated with the labeling effort.

Currently only few AL approaches~\cite{sener2018active, DBLP:conf/icml/GalIG17, DBLP:journals/corr/abs-1708-00489, DBLP:journals/tcsv/WangZLZL17, DBLP:conf/miccai/YangZCZC17} evaluated on CNNs exist. Furthermore, most of the proposed AL methods in computer vision problems are focusing on image-level classification tasks~\cite{sener2018active, DBLP:conf/cvpr/VijayanarasimhanG09,DBLP:conf/icml/KrishnamurthyAH17, DBLP:conf/icml/GalIG17,DBLP:journals/corr/abs-1708-00489, DBLP:journals/tcsv/WangZLZL17}. In contrast, very few works apply AL on CNNs with spatial and dense output spaces~\cite{DBLP:conf/miccai/YangZCZC17,DBLP:journals/corr/abs-1711-09168}.

Regarding semantic segmentation the relationship between annotation time, amount of label noise and a resulting deep model's performance with respect to its capacity has not been investigated to the best of our knowledge. Therefore, we focus our work on how to cost-effectively create \textit{reliable} training data for learning high performing CNNs for semantic segmentation by utilizing AL. We decided to use AL since, as far as we are aware of, it is the only paradigm optimized for cost-effectively generating reliable dense semantic segmentation annotations of real-world imagery data without the need of other input modalities.

In this work, we propose a novel cost-effective active learning framework tailored to multi-class semantic segmentation (\textit{CEREALS}). In particular, we aim to iteratively find the {\it minimal} set of highly informative data while {\it minimizing} the annotation effort, in order to achieve a desired high quality performance with minimal costs. We approximate costs by the amount of user interactions measured by the number of clicks performed during the annotation process. The proposed framework reduces the labeling effort by (i) utilizing spatial estimates about annotation costs inferred from a learned cost prediction CNN and (ii) by focusing on image regions promising high information content and low annotation costs in a global context. We demonstrate the performance of \textit{CEREALS} on Cityscapes~\cite{DBLP:conf/cvpr/CordtsORREBFRS16}, a very complex dataset consisting of high definition natural urban scene images. A qualitative result of our approach is depicted in Fig.\ref{fig:teaser}.\label{sec:intro}
\section{Related Work}
 
Densely annotating images with pixel-accurate multi-class semantic segmentation masks requires considerably more time than annotating images with univariate multi-class annotations~\cite{DBLP:conf/cvpr/XieKSG16}. To lessen the burden of manual annotation, an array of six different strategies has been explored in the literature. 

{\bf1) Pre-training} is a standard practice whenever the amount of available ground truth data is relatively scarce. Initially training a deep CNN on less complex but large-scale databases such as Imagenet~\cite{DBLP:conf/cvpr/DengDSLL009} results in better discrimination ability of the final model~\cite{DBLP:conf/cvpr/GirshickDDM14, DBLP:conf/cvpr/OquabBLS14, DBLP:journals/corr/HuhAE16, DBLP:conf/iccv/SunSSG17}.

{\bf2) Weakly-supervised learning} has shown promising results towards solving semantic  segmentation tasks~\cite{DBLP:conf/iccv/PapandreouCMY15, DBLP:conf/cvpr/ZhangZWX15, DBLP:conf/eccv/KolesnikovL16, DBLP:conf/eccv/ShimodaY16, DBLP:conf/eccv/SalehASPGA16, DBLP:conf/iccv/SalehASPA17, DBLP:journals/pami/LuFXHWG17, DBLP:conf/cvpr/HongYKLH17, DBLP:conf/cvpr/KhorevaBH0S17,DBLP:conf/cvpr/OhBKAFS17}. For instance, annotating a dataset just based on the annotator's binary decision if a class is present in an image or not, or annotating bounding boxes is faster than producing dense segmentation masks. However, given a sufficient amount of data, models trained in a supervised way outperform any weakly-supervised method.

{\bf3) Semi-supervised learning} methods have been shown to increase a model's performance using a mixture of fine-grained labels plus additional unlabeled data to achieve better results than labeled data alone~\cite{DBLP:conf/iccv/PapandreouCMY15,8237868,DBLP:journals/corr/abs-1802-07934}.

{\bf4) Synthetic data generation} methods produce synthetic images together with highly accurate dense annotations~\cite{DBLP:conf/eccv/RichterVRK16, DBLP:conf/bmvc/ShafaeiLS16, DBLP:journals/corr/abs-1708-01566}, but the shortcoming lies in the effort required to generate diverse and realistic sceneries~\cite{DBLP:journals/corr/abs-1708-01566}, which is a crucial aspect for achieving satisfactory results in real-world scenarios~\cite{DBLP:conf/eccv/Movshovitz-Attias16, DBLP:conf/cvpr/ZhangSYSLJF17}. 

{\bf5) Interactive segmentation} is the process of extracting objects of interest by utilizing sparse user input. It directly improves the annotation tools for assisting human annotators by increasing their efficiency~\cite{DBLP:conf/bmvc/FathiBRR11, DBLP:conf/cvpr/LuBSW16, DBLP:conf/cvpr/FengPCC16}. Though it has been recently shown that by utilizing CNNs the segmentation quality can be drastically improved compared to the previous state of the art~\cite{DBLP:conf/cvpr/XuPCYH16}, these methods are still suffering from imprecision~\cite{DBLP:conf/cvpr/CastrejonKUF17}. Castrejon \etal show how this problem could be treated by allowing the annotator to refine estimated polygons for cost-effectively generating reliable instance annotations~\cite{DBLP:conf/cvpr/CastrejonKUF17}, however the work doesn't show quantitative results on dense semantic segmentation. Xie \etal \cite{DBLP:conf/cvpr/XieKSG16} and Liang-Chieh \etal \cite{DBLP:conf/cvpr/ChenFU14} presented methods for effectively generating multitude of ground truth instance segmentations. Both methods are dependent on the utilization of lidar sensors.

{\bf6) Active learning} is described in the survey from Settles~\cite{settles2009active}. It offers a high-level overview over the commonly used methods. \textit{Pool-based active learning}~\cite{DBLP:journals/sigir/Lewis95a} exploits the inequality of amount of information in an existing unlabeled pool and aims to find the most valuable sample to be labeled by an oracle able to reveal the ground truth semantics of interest given some data. {\it AL on Semantic Segmentation} has been investigated in~\cite{DBLP:conf/cvpr/VezhnevetsBF12, DBLP:conf/iccv/KonyushkovaSF15, DBLP:conf/cvpr/Mosinska-Domanska16, DBLP:conf/cvpr/JainG16}. Both methods~\cite{DBLP:conf/cvpr/VezhnevetsBF12, DBLP:conf/iccv/KonyushkovaSF15} rely on a previously processed oversegmentation of an unlabeled image for retrieving its superpixels. Informative regions however are not restricted to the extent of superpixels. Annotating them furthermore does not guarantee a reliable labeling because oversegmentation algorithms often fail to separate semantic regions when the transition from one class to the next is smooth in the underlying input space. The method proposed by Mosinska-Domanska \etal \cite{DBLP:conf/cvpr/Mosinska-Domanska16} operates only on curvilinear structures and \cite{DBLP:conf/cvpr/JainG16} is investigating the propagation of segmentations from informative data to unlabeled data. {\it AL for CNNs on Semantic Segmentation} running on top of unstructured state-of-the-art models, namely fully convolutional neural networks~\cite{DBLP:conf/cvpr/LongSD15}, has so far to the best of our knowledge only been investigated in~\cite{DBLP:conf/miccai/YangZCZC17,DBLP:journals/corr/abs-1711-09168}. Both approaches focus on foreground/background segmentation and assume an equal annotation effort for all images, which we later show to be a simplified assumption. {\it Cost-Effective AL for CNNs} has been recently proposed in~\cite{DBLP:journals/tcsv/WangZLZL17} for image classification tasks, where highest confidence pseudo-annotated unlabeled samples are added to the training set with no human cost at all. The same idea has been adopted in~\cite{DBLP:journals/corr/abs-1711-09168} for medical image segmentation. This method assumes that highly confident predictions are labeled correctly; Hence, selected samples could introduce hard label noise during training. Although this technique shows an improvement regarding general performance, it may lead to unwanted side-effects on corner-cases due to strengthening wrong, but certain predictions. Wherever in this work the authors propose to score the entire images, we are 1.) scoring all possible fixed-size regions across all images in an unlabeled pool and 2.) querying the ones for annotation expected to have the highest positive impact on the model's performance. Most related to our approach regarding cost-effectiveness in active learning are the methods described in~\cite{Settles_activelearning, DBLP:conf/cvpr/VijayanarasimhanG09, DBLP:conf/icml/KrishnamurthyAH17}. All these works employ a cost prediction model trained on data where target labels are available from the beginning or after previously executed acquisition steps. We adapt this idea to the domain of semantic segmentation by estimating spatial information about costs in order to find highly informative, but cheap regions in an unlabeled pool of images to be annotated, where the quality of annotation depends only on the quality of the executing annotators and their given labeling instructions. Despite the fact that AL could be incorporated alongside all the previously mentioned approaches, in this work we are considering a CNN, specifically a fully convolutional neural network, trained in a strongly supervised manner and a simple polygon-based annotation tool, similar to ones used for constructing training datasets of state-of-the-art benchmarks~\cite{DBLP:conf/cvpr/CordtsORREBFRS16,DBLP:conf/iccv/NeuholdOBK17,DBLP:conf/cvpr/ZhouZPFB017}. Though not being optimized for efficiency, it allows the production of fine-grained annotations~\cite{DBLP:journals/ijcv/RussellTMF08} for training high quality CNNs on multi-class semantic segmentation tasks.
\label{sec:related}
\section{Method}

In classical pool-based AL typically only a single sample out of an unlabeled pool is queried to be labeled by an oracle in each step of the iterative algorithm. Since deep CNN training algorithms need a long time to converge on currently available hardware, such a setup is however practically infeasible. Therefore, we consider a pool-based AL scenario running in batch-mode. In such a setting a large unlabeled pool of data exists from which a small, randomly sampled subset, called the seed set, is initially extracted and labeled by an oracle. Using this seed set the algorithm works as follows: First, a model is trained on the currently labeled pool. Secondly, some measure of information on each individual unlabeled sample is being computed. Thirdly, an acquisition function is applied. A subset of a pre-specified amount of elements maximizing the acquisition function is annotated by an oracle. It is then added to the labeled pool. The process is repeated until either a desired performance or labeling budget is reached. Furthermore, the stopping criterion is satisfied whenever the unlabeled pool becomes exhausted which is indicated by no further improvements after several acquisition steps.

\begin{figure}[!ht]{\includegraphics[width=1\textwidth]{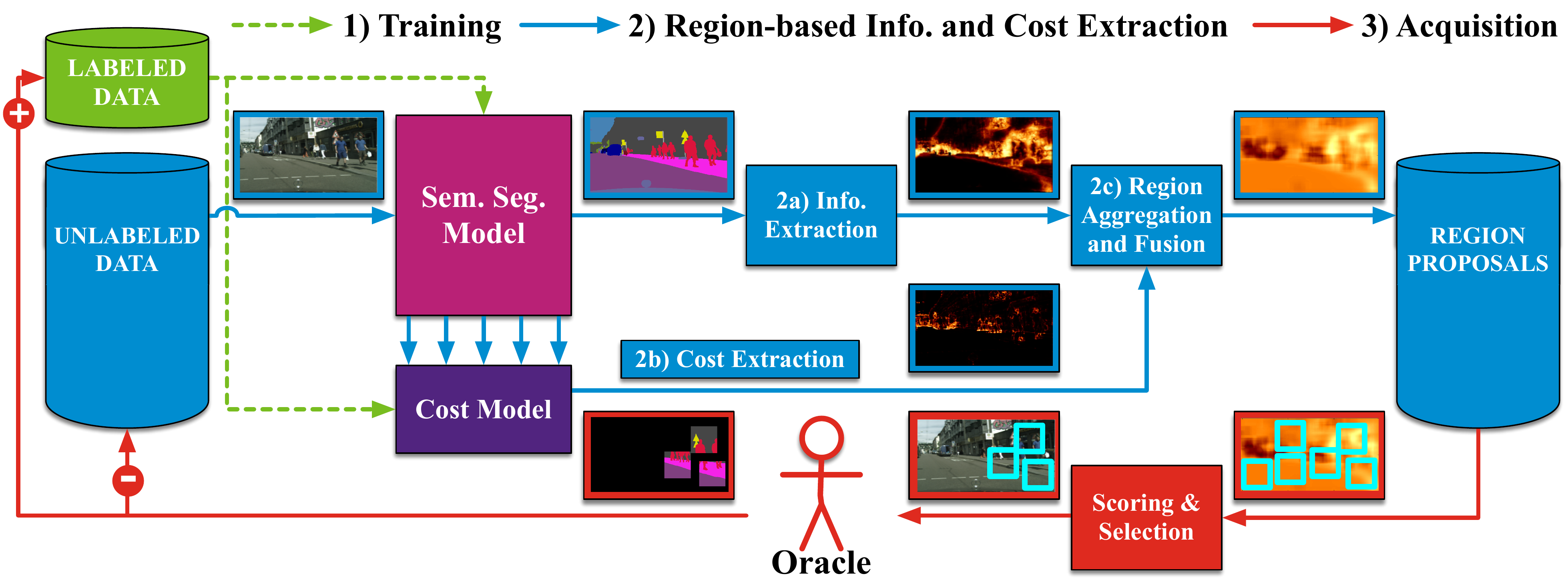}}
\centering
\caption{Diagram of the proposed framework for cost effective region-based active learning for semantic segmentation. We fuse spatial information about information content and cost estimates in order to query the most promising regions for annotation maximizing an information/cost trade-off. Our approach consequently aims to boost the performance of a CNN as cheaply as possible.}
\label{fig:method:cereals}
\vspace{-0.25cm}
\end{figure}

The main attention in pool-based AL research has been given to information measures being computed on the posterior probability distribution of a classifier given some input data. An acquisition function for batch-mode AL scenarios is often extended with density weighting approaches aiming to select samples maximizing not only information content but also diversity. Since the problem of semantic segmentation does not only allow the exploration of novel information measures and/or diversity maximization methods, our approach focuses on the acquisition process itself. In a typical AL scenario for image classification problems, a subset of promising images from an unlabeled pool is sampled. Such an acquisition function has been adopted for semantic segmentation in ~\cite{DBLP:journals/corr/abs-1711-09168} for retrieving images, but based on their accumulated per-pixel information content projecting all the information extracted from an image onto a single value. We are proposing to design acquisition to explicitly focus on image regions inside of the entire unlabeled pool of images and further to not only consider information during region selection but also annotation costs. The proposed method is depicted in Fig.\ref{fig:method:cereals} and works as follows:

\vspace{-0.25cm}
\paragraph{1) Training}
\label{method:training}
For constructing the seed set we uniformly sample $n$ images to be fully labeled by an oracle. We then learn two deep convolutional neural networks. One being the \textit{semantic segmentation model} based on the \textit{FCN8s} architecture~\cite{DBLP:conf/cvpr/LongSD15}, whose training is initialized using imagenet-pretrained weights. For faster training we apply the width multiplier introduced in \cite{DBLP:journals/corr/HowardZCKWWAA17} with its value set to $0.25$. Furthermore, we discarded the $8\times$ upsampling and instead scale down the spatial annotations during training, since we observe it to have only marginal impact on the model's final performance. For validation however, we add a $8\times$ bilinear upsampling layer. The other model, which is trained directly after the \textit{semantic segmentation model}, is a \textit{cost model} based on~\cite{DBLP:conf/cvpr/KirillovLASR17}. It utilizes the semantic segmentation networks learned knowledge as prior information to estimate the clicks an annotator would have needed to execute for densely annotating an image. All further implementation details are reported in \ref{appendix:impl_details}. 

\begin{figure}[!ht]
\begin{subfigure}[Est. Semantic Segmentation]{\includegraphics[width=0.329\textwidth]{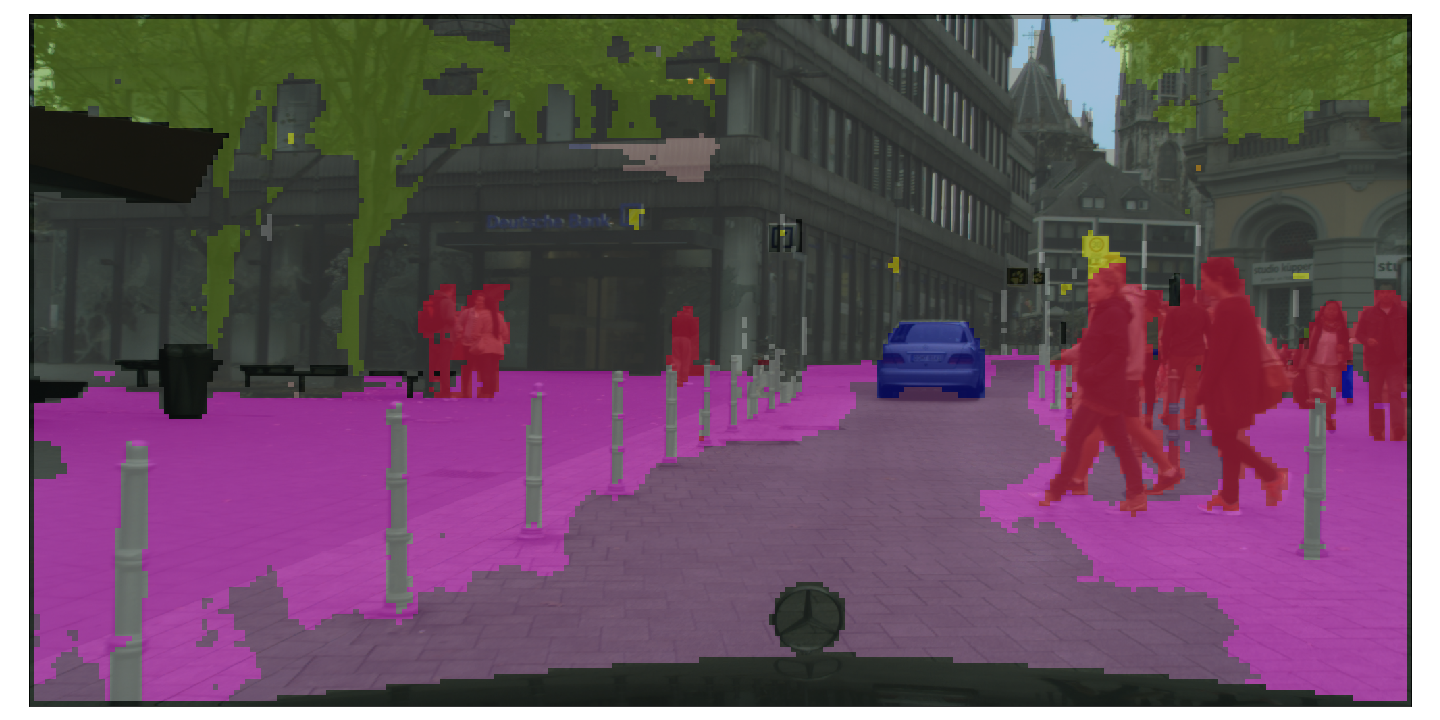}\label{fig:method:semseg}}\end{subfigure}
\begin{subfigure}[Information Map]{\includegraphics[width=0.329\textwidth]{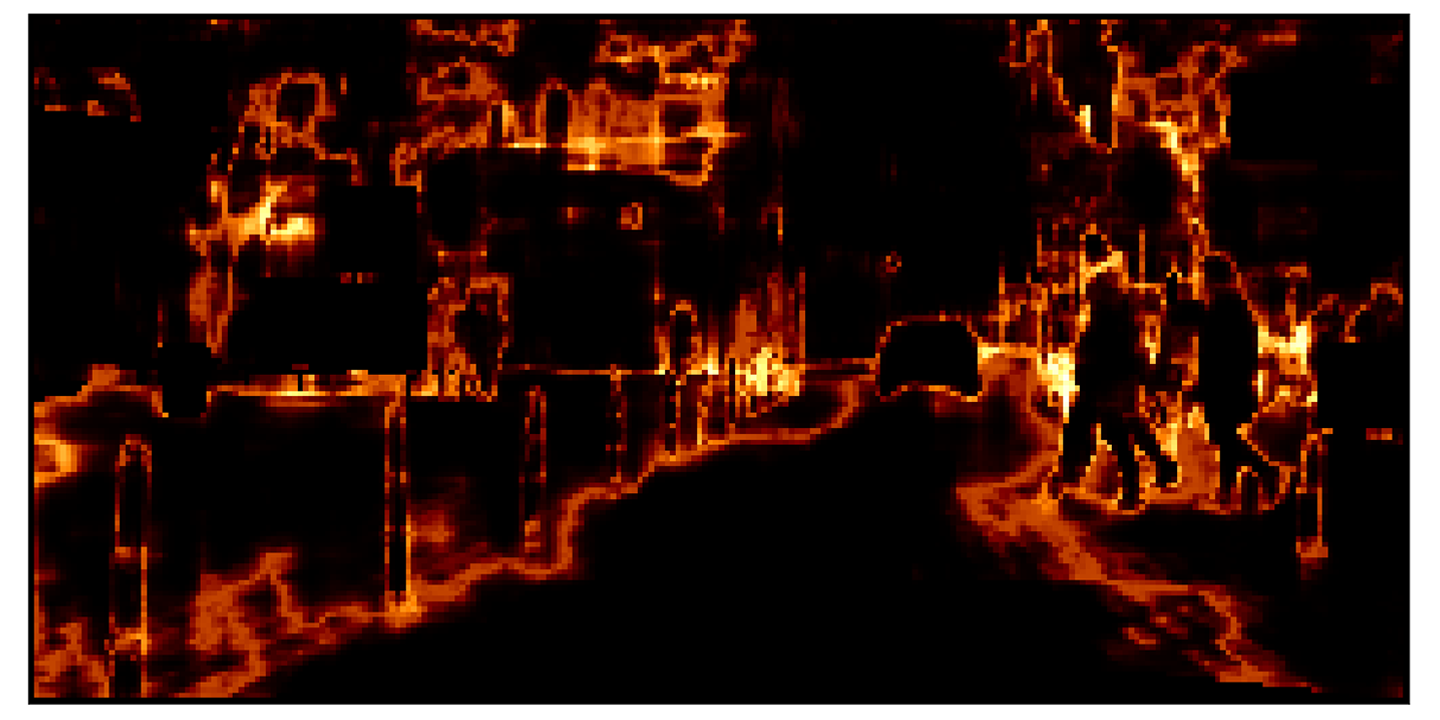}\label{fig:method:information_map}}\end{subfigure}
\begin{subfigure}[Est. Cost Map]{\includegraphics[width=0.329\textwidth]{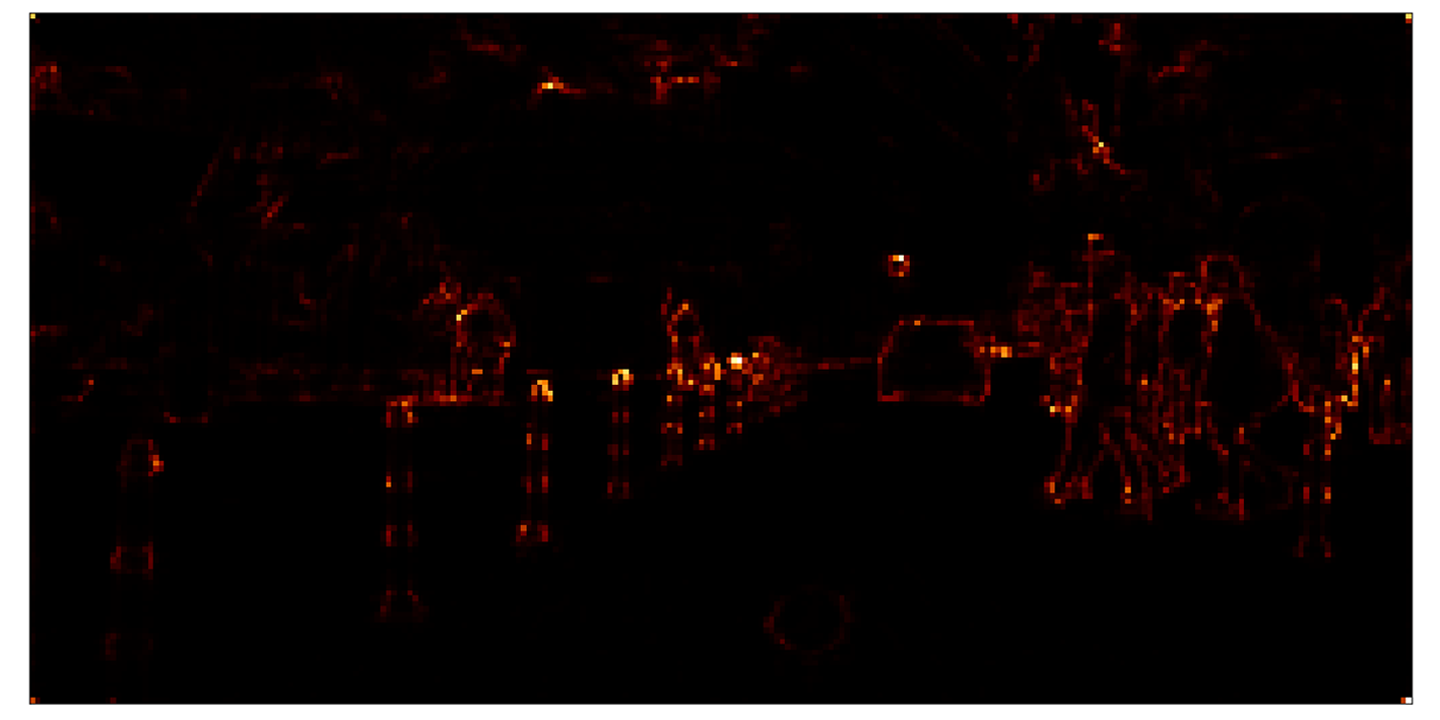}\label{fig:method:cost_map}}\end{subfigure}\\

\begin{subfigure}[Fused Region Map]{\includegraphics[width=0.329\textwidth]{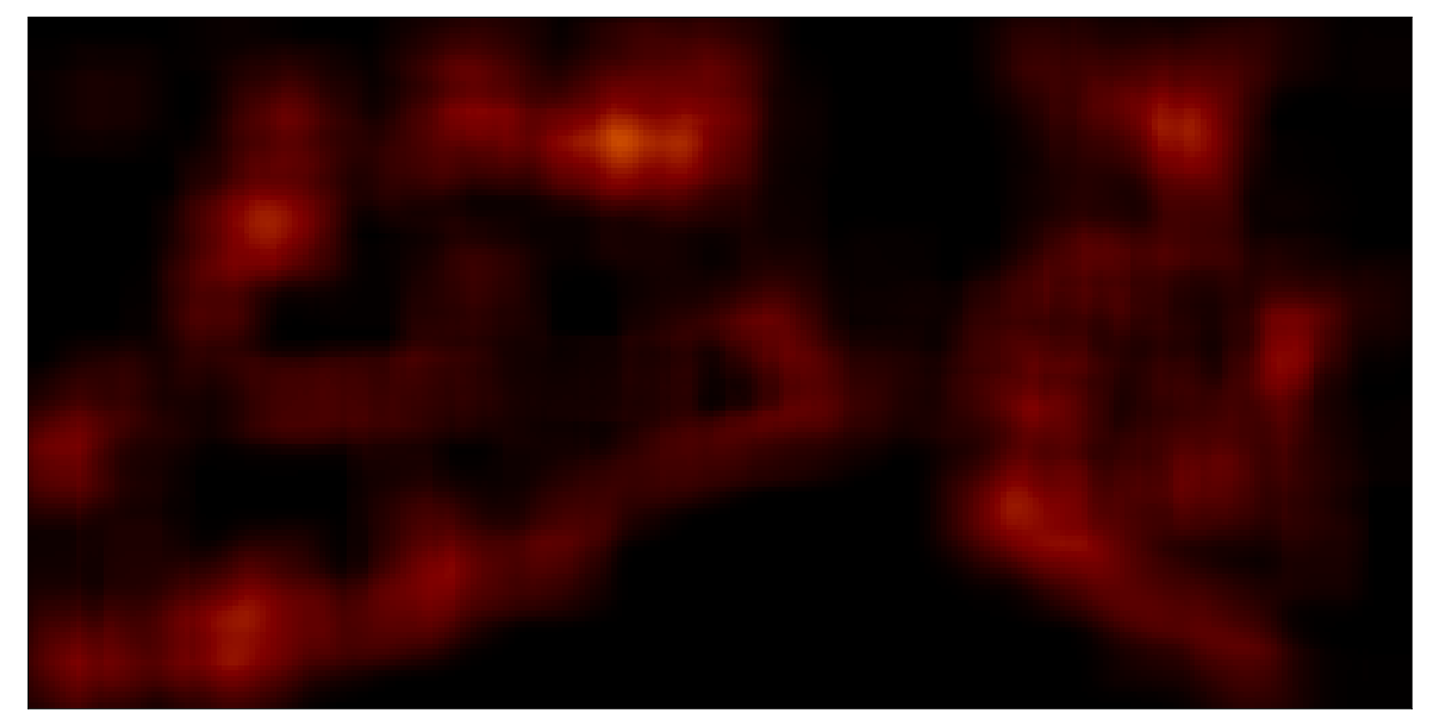}\label{fig:method:fused}}\end{subfigure}
\begin{subfigure}[Region Proposals]{\includegraphics[width=0.329\textwidth]{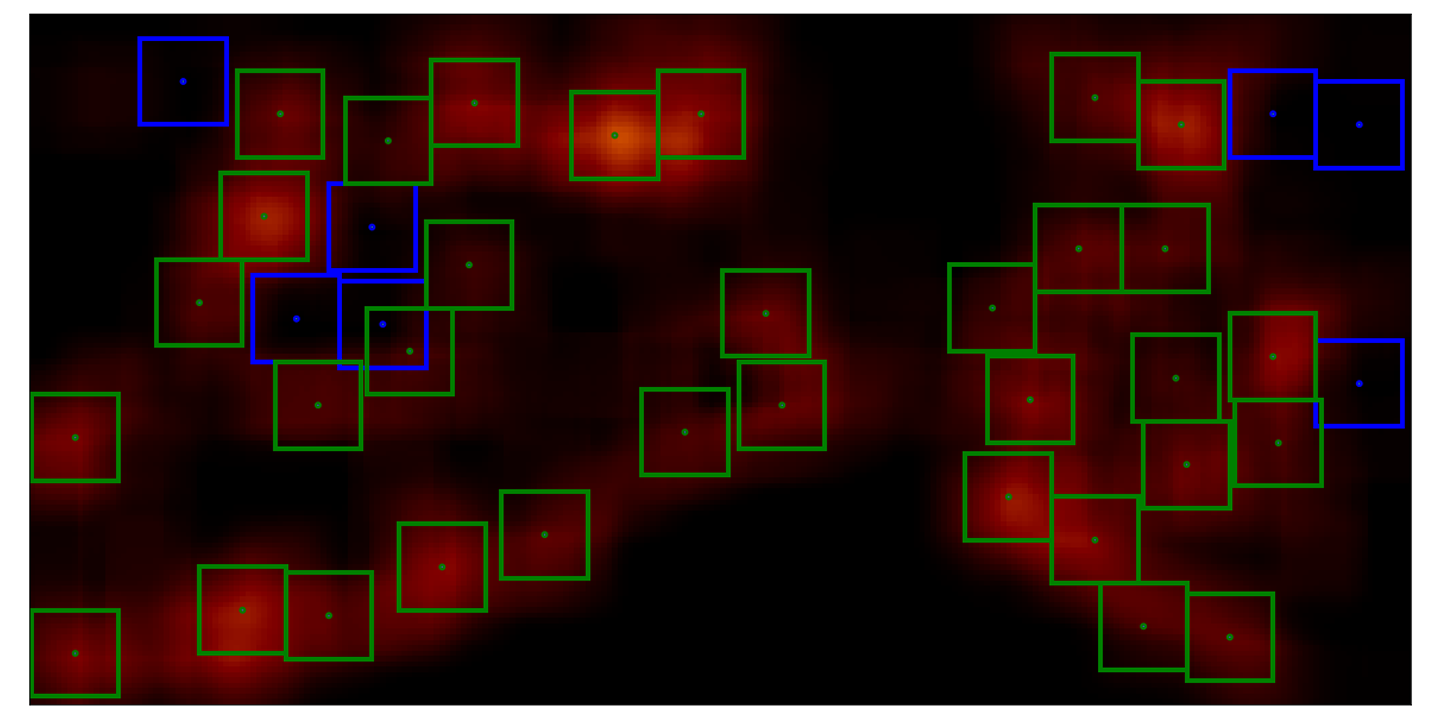}\label{fig:method:region_proposals}}\end{subfigure}
\begin{subfigure}[Annotated Regions]{\includegraphics[width=0.329\textwidth]{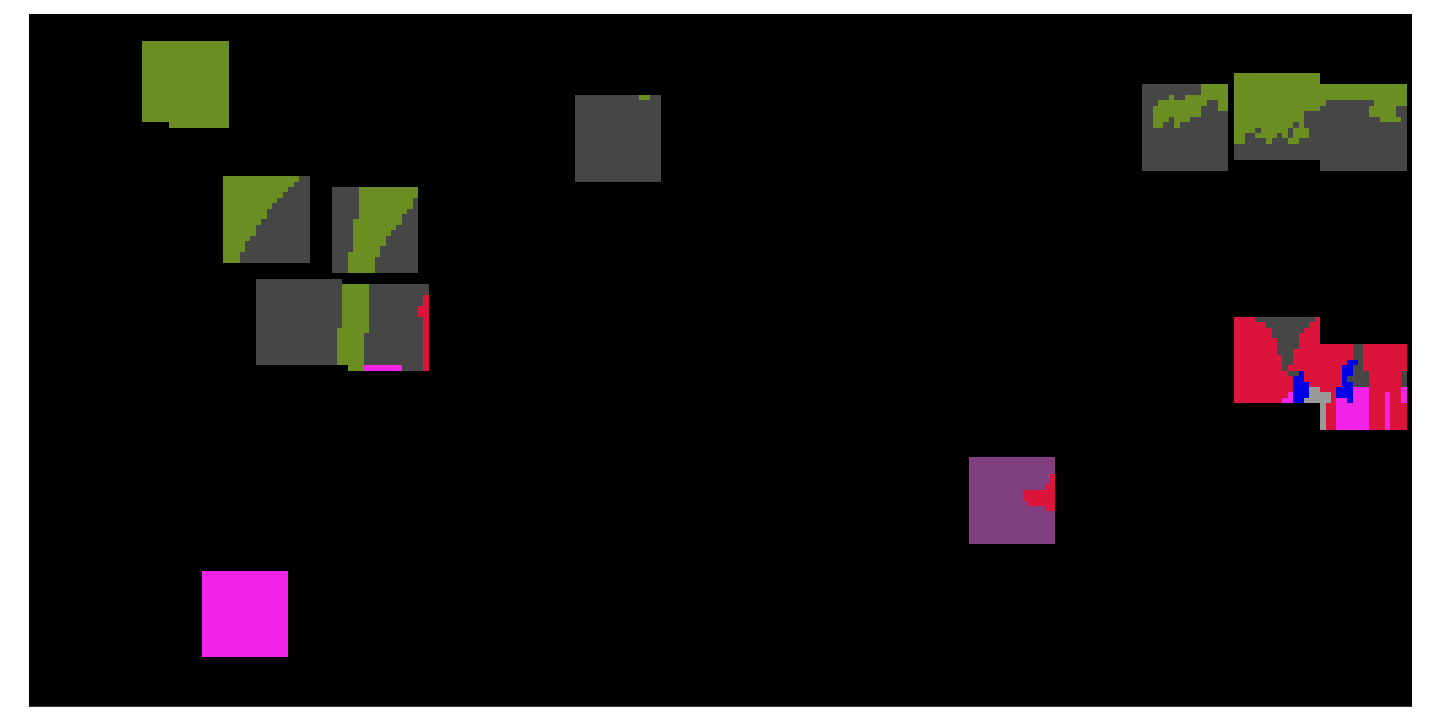}\label{fig:method:annotation}}\end{subfigure}

\caption{Visualization of \textit{CEREALS} query selection behaviour during acquisition step two. Blue boxes in (e) represent regions annotated at the end of the previous acquisition step one. Green boxes represent regions within the region proposal pool of the current acquisition step. \textit{CEREALS} queries the best regions out of the region proposal pool for annotation (f).}
\vspace{-0.25cm}
\end{figure}

\vspace{-0.275cm}
\paragraph{2a) Information Extraction}
\label{method:acquisition_functions}
In this work we raise awareness towards costs. We therefore compare two classical heterogeneous information measures only. We are computing both information measures for each pixel location individually given the {\it a-posteriori} probability distributions retrieved from the activations produced by the employed semantic segmentation CNN's softmax layer. In the following formulas $P^{(u,v)} \widehat{=} P^{(u,v)}(f_{\theta}(x))$ is the probability class distribution at a specific pixel position $(u,v)$ retrieved from a model $f$ parameterized by $\theta$ given some image $x$. A specific class out of a set of considered classes is denoted by $c$. The resulting {\it information map} (Fig.\ref{fig:method:information_map}) contains the information content for each pixel of an image at a current acquisition step.  

Entropy~\cite{DBLP:journals/sigmobile/Shannon01} is the most widely used information measure seen in active learning literature. Here, the data with the highest positive impact on the model's performance is estimated to be the one where it's posterior probability distribution produces the highest entropy. Entropy is, inter alia, used as a measure of uncertainty, since its value is maximized when the model assigns each considered class the same probability and very small if the model is sure about its decision. We compute entropy for retrieving per-pixel information as follows.

\begin{equation}
  \label{eq2}
  H^{(u,v)} := -\sum\limits_{c} P_c^{(u,v)} \cdot log(P_c^{(u,v)}) 
\end{equation}

The Vote Entropy~\cite{DBLP:conf/icml/DaganE95} information measure entails first constructing a committee $E$ of $N_E$ different classifiers that ideally are all consistent with the labeled pool. Each committee member $e$ places a vote on vector $P_e^{{(u,v)}}$. Then a disagreement factor among the members is calculated. We utilize vote entropy which we adapt for the semantic segmentation case as follows. 
\vspace{-0.5cm}

\begin{equation}
\mbox{\fontsize{10}{10}\selectfont\( %
  \label{eq3}
  {V}^{(u,v)} := -\sum\limits_{c} \frac{\sum\limits_{e} D(P_{e}^{{(u,v)}},c)}{N_E} \cdot log \frac{\sum\limits_{e} D(P_{e}^{{(u,v)}},c)}{N_E}
  \quad where \quad
  D(a,c)=
\begin{cases}
    1, & \text{if } \argmax(a)=c\\
    0, & \text{otherwise}
\end{cases}
\)} %
\end{equation}

Instead of training $N_E$ different classifiers on the same training data, we leverage the stochasticity provided by the dropout layers of our employed \textit{semantic segmentation model} and construct a Monte-Carlo dropout ensemble as in \cite{DBLP:conf/icml/GalG16}. The most informative data points are the ones having the highest disagreement factor among the committee members. The aim of such \textit{Query by Committee}~\cite{DBLP:conf/nips/FreundSST92} approaches is to sample data expected to reduce the version space of given committee members. 

\vspace{-0.275cm}
\paragraph{2b) Cost Extraction}
Our work is based on the assumption that some samples in an unlabeled pool are more costly to label by a human oracle than others and further that this also applies to regions within images. To the best of our knowledge no published dataset addressing semantic segmentation provides information on annotation costs. As in~\cite{DBLP:conf/cvpr/FengPCC16,DBLP:conf/cvpr/XuPCYH16}, we are approximating costs by the number of clicks necessary to annotate an image. Cityscapes is the only dataset providing information about where and how often a user has clicked to label an image. Obviously, this information is unknown for unlabeled data. For this reason we train the \textit{cost model} on all the click data which was produced by human annotators at previous acquisition steps. During actual cost extraction we perform a forward pass for each individual image within the current unlabeled pool through the \textit{cost model} for retrieving an estimate about clicks. We denote the result given an image as {\it cost map} (Fig.\ref{fig:method:cost_map}).

\vspace{-0.275cm}
\paragraph{2c) Region Aggregation and Fusion}
We argue that not all regions in an image boost a CNN's performance equally. Thus, some regions may have not only different labeling costs but also may provide supervisory signals of different impact on the model's performance than other regions. Theoretically, some regions could be very costly to label while having only little positive impact on the models performance and vice-versa. Based on this assumption, we leverage the varying information content and cost of regions within an unlabeled pool of images in order to query the highest density samples to be passed to a human oracle for labeling. We aim to maximize a trade-off, such that for the minimal cost we achieve maximal performance. Various definitions of regions exist. In this work, we only consider regions of quadratic shape and investigate the impact on the employed model's performance regarding their varying sizes. 

We utilize a sliding-window approach for selecting the most informative regions from within the acquired {\it information maps} computed for each individual image of the unlabeled pool. For a pre-specified window size we proceed as follows: At each sliding-window location $(u,v)$, we accumulate all the values of our {\it information map} encompassing the dimensions of the window and store this density in a matrix denoted as {\it region information map} having the same spatial dimensionality as the considered image. We proceed similarly to generate {\it region cost maps} given the estimated {\it cost maps}. 

We linearly scale {\it region information maps} and {\it region cost maps} w.r.t. the whole dataset, such that all values are in $[0, 1]$. We then fuse corresponding region maps using one of the following fusion functions. The three simple fusion functions we have evaluated are denoted in \eqref{eq6}, \eqref{eq7}, and \eqref{eq8} with the \textit{region information map} $I$ and the \textit{region cost map} $C$. The parameter $\alpha$ in \eqref{eq8} allows to set a trade-off for linearly interpolating between both region maps. An example of a resulting \textit{fused region map} is depicted in Fig.\ref{fig:method:fused}.  
\vspace{-0.25cm}

\begin{tabular}{p{0.19\textwidth}p{0.25\textwidth}p{0.38\textwidth}}
{\begin{align}
\label{eq6}
g_1=\frac{I}{1+C}
\end{align}}
&
{\begin{align}
\label{eq7}
g_2=(1-C) \cdot I
\end{align}}
&
{\begin{align}
\label{eq8}
g_3=I \cdot\alpha+(1-C) \cdot (1-\alpha)
\end{align}}
\end{tabular}

\vspace{-0.25cm}
After fusing the region information and the region cost map pairs for all images in the current unlabeled pool, we perform non-maximum-suppression to retrieve fixed-size region candidates for each individual image. Regarding non-maximum-suppression we always favor higher scoring regions regarding its computed information/cost trade-off while not allowing any overlap until maximum coverage. We store the region candidates (Fig.\ref{fig:method:region_proposals}) for each individual image of the unlabeled pool within a region proposal pool. Note that the region candidates are not allowed to overlap in between a round, since in an asynchronous annotation mode we do not want to assign the same pixels to be labeled to different annotators. 

\vspace{-0.275cm}
\paragraph{3) Acquisition} From the region proposal pool we extract as many top scoring regions as would correspond to extracting $m$ images out of a pool of equally sized images regarding their amount of pixels for a fair comparison to the image-based acquisition of labels (Fig.\ref{fig:method:annotation}). Instead of employing a real annotator for evaluating our method we utilize a robot user as our oracle. Whenever annotations are being requested, the robot user uses the ground truth annotation of the considered training set. We then update the labeled and unlabeled pool and learn our \textit{semantic segmentation model} and \textit{cost model} from scratch.  

\label{sec:method}
\begin{figure}[!ht]
\begin{subfigure}[]{\includegraphics[trim={0.25cm 0 1.8cm 0},clip,width=0.495\textwidth]{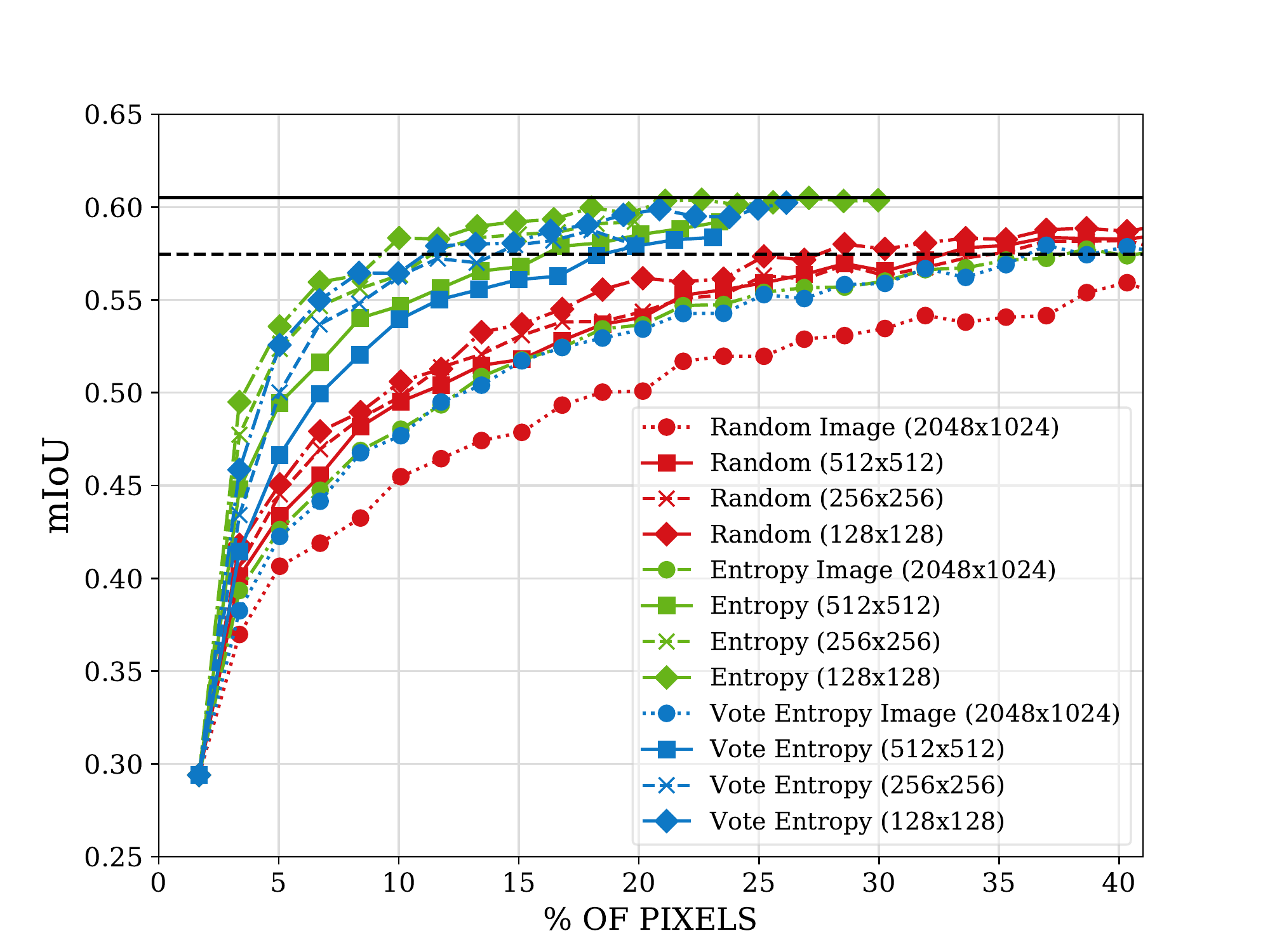}\label{fig:results:a}}\end{subfigure}
\begin{subfigure}[]{\includegraphics[trim={0.25cm 0 1.8cm 0},clip,width=0.495\textwidth]{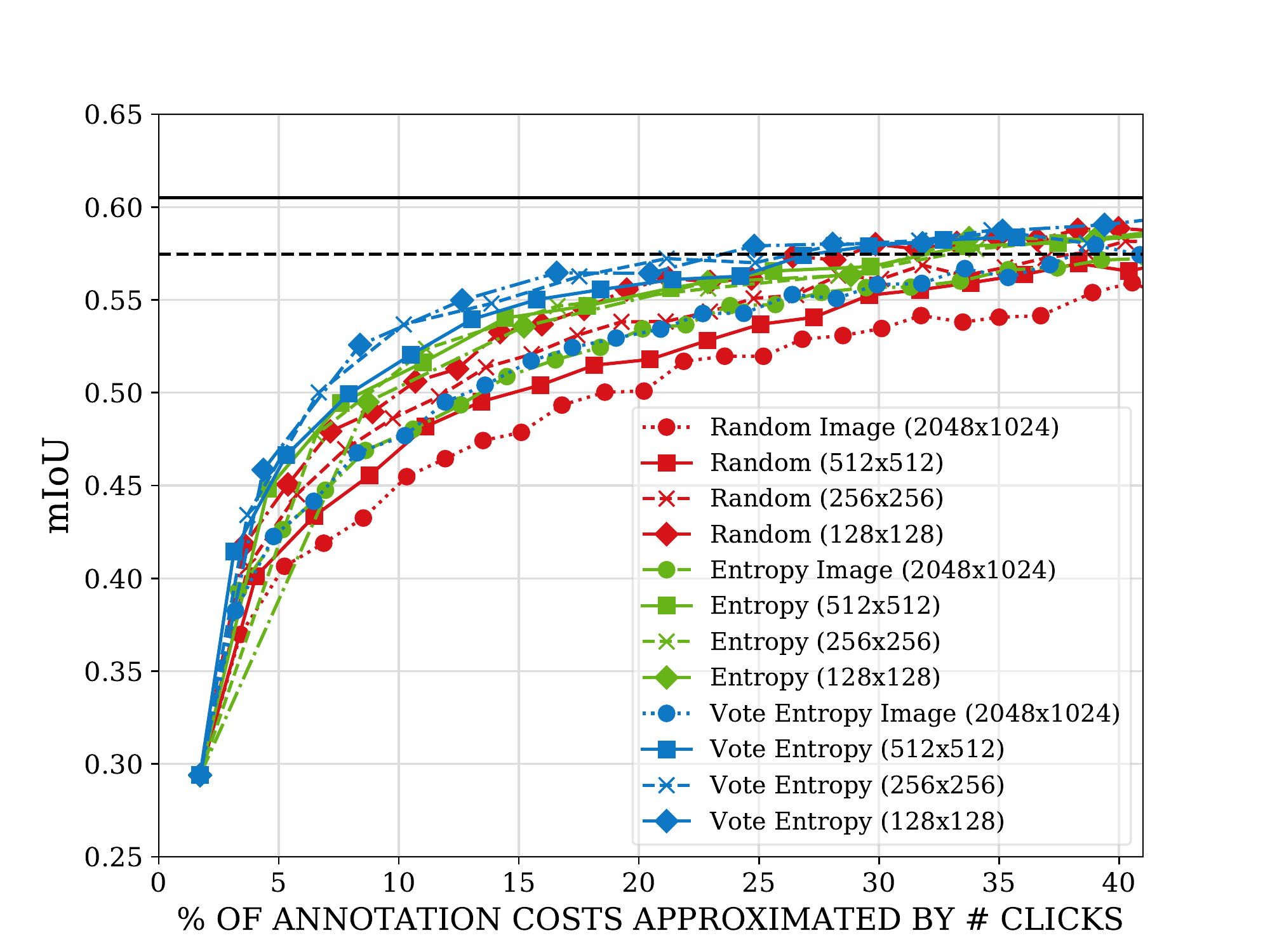}\label{fig:results:b}}\end{subfigure}
\caption{AL curves showing the relationship between pixels and annotation costs approximated by the number of clicks regarding different acquisition functions. The solid black line shows the mIoU achieved by training the model on the whole training set of Cityscapes. The dashed black line marks $95\%$ of the performance achieved by this model. a) Resulting mIoU as function over the amount of labeled pixels queried from an annotator. b) Same obtained results but plotted as a function over the annotation effort measured by the number of clicks. }
\end{figure}

\section{Results}
All processed experiments presented in this work are repeated five times and we report the average \textit{mean Intersection over Union} (mIoU) calculated on the validation dataset of Cityscapes after training convergences. We claim convergence whenever a model's mIoU on the validation dataset does not increase within ten epochs.
The seed set is initialized to $n=50$ fully annotated images randomly selected from the unlabeled pool. 

Our experiments are structured as follows: First, we explore the impact of varying region sizes on the models performance w.r.t. the number of queried pixels. Secondly, we show how the results relate w.r.t the labeling costs approximated by the number of clicks. Thirdly, we provide evidence that knowledge about costs can be utilized to further reduce labeling efforts.

Finally, we demonstrate that knowledge about costs can be inferred from a learned CNN, regressing spatial information about costs. 

In our first experiment, we are querying the $m=50$ top scoring images maximizing the considered information measures only, exactly as suggested in \cite{DBLP:journals/corr/abs-1711-09168}. We do not perform any special treatment of semantic boundaries.

\begin{figure}[!ht]
\centering
\begin{subfigure}[]{\includegraphics[trim={0.25cm 0 1.8cm 0},clip,width=0.495\linewidth]{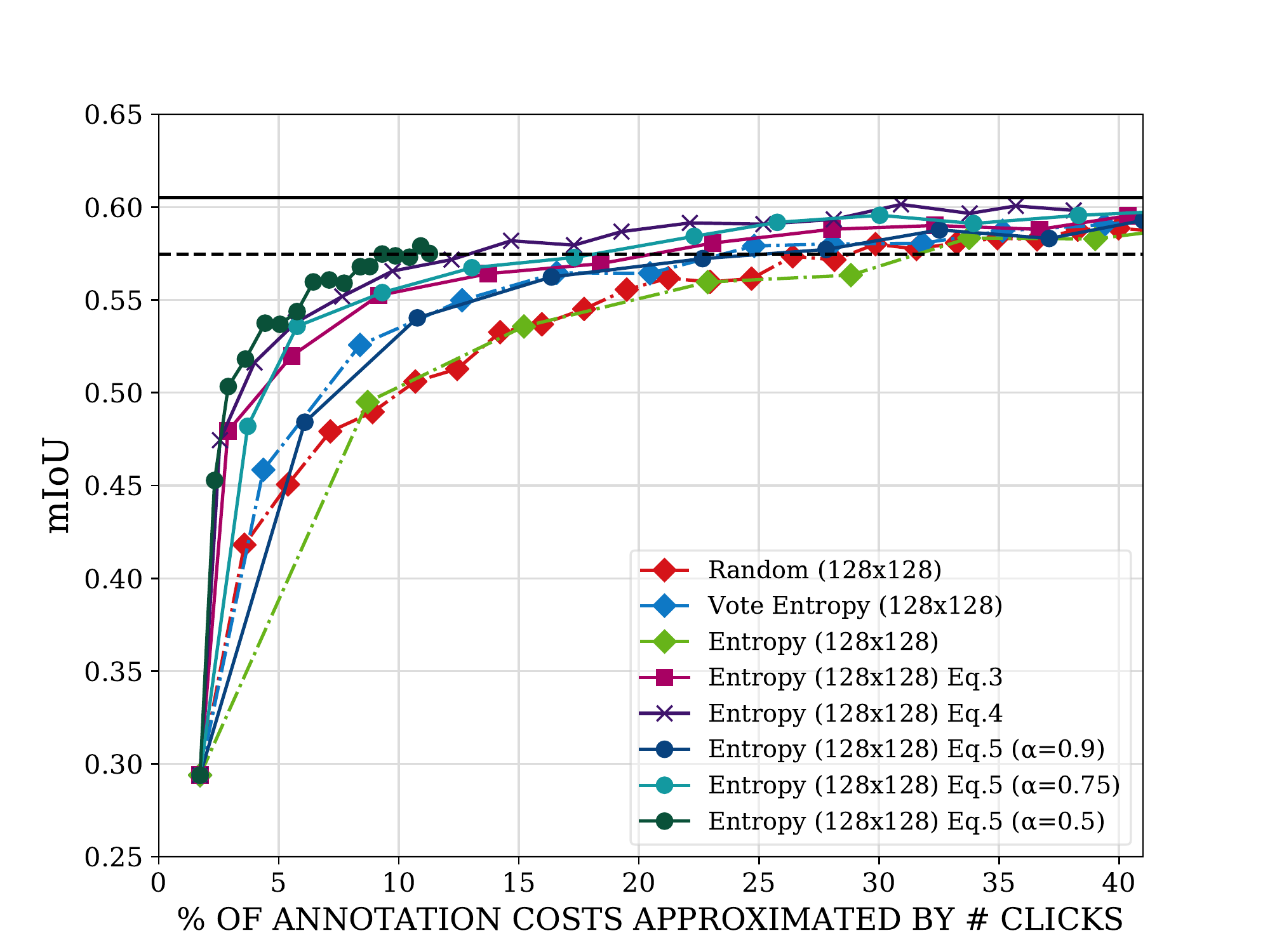}\label{fig:results:costs}}\end{subfigure}
\begin{subfigure}[]{\includegraphics[trim={0.25cm 0 1.8cm 0},clip,width=0.495\linewidth]{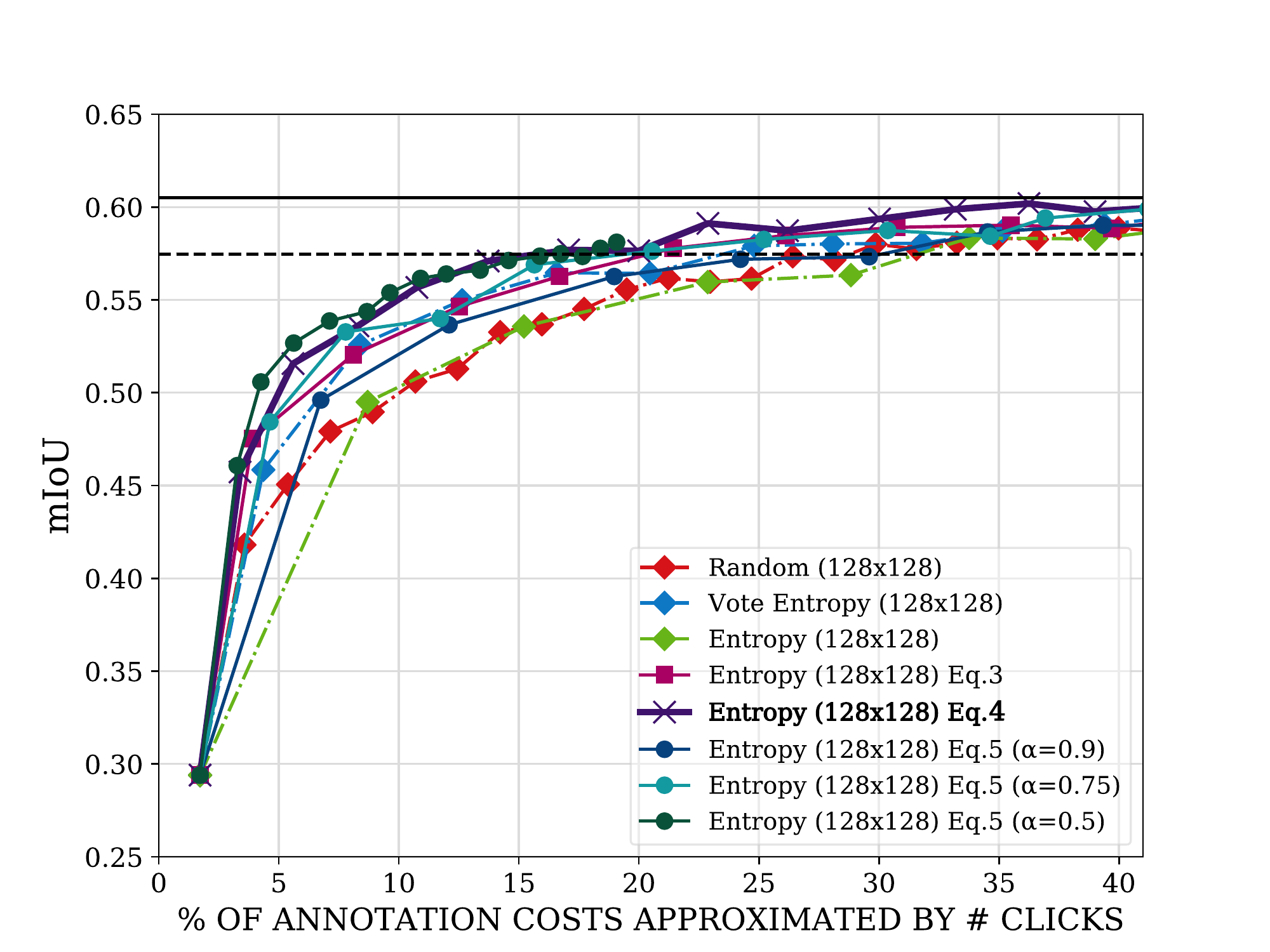}\label{fig:results:estimated_costs}}\end{subfigure}\\
\caption{In this plot we show the results achieved by our region-based acquisition function when optimizing towards both; minimal costs and high information using entropy sampling. a) Selecting regions by using the ground truth clicks, in order to show how our method would perform if the utilized cost prediction network would always perform perfectly b) Selection of regions by entropy and minimization of estimated costs.}
\vspace{-0.4cm}
\end{figure}

In Fig.\ref{fig:results:a} we plot the obtained results regarding the percentage of labeled pixels relative to all labels present in the training dataset of Cityscapes against the achieved mIoU of the trained CNN. All considered acquisition functions show better results than random sampling. After $21$ acquisition steps corresponding to $35.29\%$ of queried labels by using entropy sampling, we achieve an mIoU of $0.575$ which corresponds to $95\%$ of the performance as compared to the obtained result of $0.605$, when training on the full training set of Cityscapes. We will refer to the former performance measure as $p95$ and to the latter as $p100$.

In our second experiment, we are evaluating the region-based acquisition of labels and sample $512\times512$-sized most promising regions out of the entire unlabeled pool. Despite the very large region size we observe a significant improvement for all evaluated information measures just caused by the spatial exploitation of the unlabeled set. We are achieving $p95=16.74\%$ by using entropy sampling. The region-based random sampling approach also shows better results compared to sampling whole images randomly, which we argue must be due to the increase in data variability introduced by sampling regions instead of entire images.
We then proceed to investigate smaller region sizes, concretely $256\times256$ and $128\times128$ and observe the performance to increase. We argue, that this is because smaller region sizes allow querying higher density regions.

An even more interesting result is found when one compares mIoU vs. the effort measured by the number of clicks relative to the total number of clicks which were executed to annotate the whole training dataset of Cityscapes. Similarly to $p95$, $c95$ will denote the performance index for achieving $95\%$ of the performance related to the amount of clicks relative to $c100$ standing for the performance which was achieved with all polygon base points of the training set. Note that $p100=c100$ when training on whole images. During evaluation we also count every additional click that might occur on the region borders for a fair comparison, such that theoretically the value of $c100$ could get bigger than $p100$ when sampling regions. The results indicate that highly informative data is also costly to label (Fig.\ref{fig:results:b}). For sampling whole images based on the entropy information measure, where $p95=35.29\%$ we see that this corresponds to $c95=39.2\%$. Furthermore the gain in using a sophisticated information measure compared to random sampling regarding the effort is much smaller when sampling regions than initially indicated when comparing against the amount of pixels. This can be clearly seen by comparing  Fig.\ref{fig:results:a} with Fig.\ref{fig:results:b}. For example, for $128\times128$ regions maximizing the acquisition function based on the entropy information measure we achieved $p95=10.01\%$ compared to $c95=33.76\%$. The queried $10.01\%$ of labels thus require an annotation effort of $33.76\%$. We report all results in \ref{appendix:results}. In terms of the best performing $128\times128$ setting the results show worse numbers for highly informative region sampling using the entropy information measure compared to the random selection of non-overlapping regions. We also observe entropy sampling to prefer more costly regions than vote entropy sampling, which is slightly better then random. 

In order to further reduce labeling effort while looking for highly informative regions we now leave the region size fixed to the best performing setting regarding our previous experiments ($128\times128$). Equations \eqref{eq6}, \eqref{eq7} and \eqref{eq8} are evaluated with different $\alpha\in\{0.5, 0.75, 0.9\}$ in order to empirically determine a good information/cost trade-off. We first establish an upper bound for the utilized information measures fused with optimal cost estimates by using the actual ground truth data about clicks. We observe that entropy sampling fused with the information about true costs to achieve better results than the more powerful vote entropy information measure alone (Fig.\ref{fig:results:costs}). We can see our method to achieve $c95=14.68\%$ using fast to compute entropy sampling assuming that the cost prediction network would always decide correctly. 

We now utilize the \textit{cost model} which is trained on the oracle feedback acquired at previous acquisition steps. With our multiplicative fusion approach we reach a performance of $c95=17.07\%$  (Fig.\ref{fig:results:estimated_costs}).

\label{sec:results}
\section{Conclusion}

We have proposed a novel method for cost effective active learning for semantic segmentation tailored to fully convolutional neural networks. We have demonstrated our framework's performance on Cityscapes, a highly diverse high definition dataset consisting of images of urban scenes captured in the wild. We show that combining information content and cost estimates is a powerful approach for cost-effectively building new training datasets from scratch. With only 17\% of the effort measured by the amount of clicks which were executed for annotating the Cityscapes training set, we are able to achieve 95\% of the full training set's performance.

We leave the question on how the performance of \textit{CEREALS} scales to other network architectures with varying representational power for further research, due to the high computational demands of such an evaluation. Furthermore, we want to encourage the community to provide ground truth information about human annotation costs of upcoming and, if available, already existing manually labeled computer vision datasets. This will help further research in cost-effectively learning high performance models in data-hungry deep learning era.

\section{Acknowledgements}
We thank the Center for Information Services and High Performance Computing (ZIH) at TU Dresden for generous allocations of computer time. Furthermore we thank our colleagues, Lucas Rego Drumond\raise0.5ex\hbox{1}, Thomas Wenzel\raise0.5ex\hbox{1}, Dimitrios Bariamis\raise0.5ex\hbox{1}, Uwe Brosch\raise0.5ex\hbox{1}, Masato Takami\raise0.5ex\hbox{1}, Alexander Lengsfeld\raise0.5ex\hbox{1}, Jens Mehnert and Volker Fischer for helpful discussions. We also thank the anonymous reviewers for their valuable comments. 

\newpage
\label{sec:conclusion}


\interlinepenalty=10000
\bibliography{bib/conferences,bib/semantic_segmentation,bib/weakly_supervised,bib/synthetic,bib/interactive_segmentation,bib/co_segmentation,bib/active_learning,bib/others}

\begin{thebibliography}{59}
\providecommand{\natexlab}[1]{#1}
\providecommand{\url}[1]{\texttt{#1}}
\expandafter\ifx\csname urlstyle\endcsname\relax
  \providecommand{\doi}[1]{doi: #1}\else
  \providecommand{\doi}{doi: \begingroup \urlstyle{rm}\Url}\fi

\bibitem[Alhaija et~al.(2017)Alhaija, Mustikovela, Mescheder, Geiger, and
  Rother]{DBLP:journals/corr/abs-1708-01566}
Hassan~Abu Alhaija, Siva~Karthik Mustikovela, Lars~M. Mescheder, Andreas
  Geiger, and Carsten Rother.
\newblock Augmented reality meets computer vision : Efficient data generation
  for urban driving scenes.
\newblock \emph{CoRR}, abs/1708.01566, 2017.
\newblock URL \url{http://arxiv.org/abs/1708.01566}.

\bibitem[Castrejon et~al.(2017)Castrejon, Kundu, Urtasun, and
  Fidler]{DBLP:conf/cvpr/CastrejonKUF17}
Lluis Castrejon, Kaustav Kundu, Raquel Urtasun, and Sanja Fidler.
\newblock Annotating object instances with a polygon-rnn.
\newblock In \emph{2017 {IEEE} Conference on Computer Vision and Pattern
  Recognition, {CVPR} 2017, Honolulu, HI, USA, July 21-26, 2017}, pages
  4485--4493, 2017.
\newblock \doi{10.1109/CVPR.2017.477}.
\newblock URL \url{https://doi.org/10.1109/CVPR.2017.477}.

\bibitem[Chen et~al.(2014)Chen, Fidler, and Urtasun]{DBLP:conf/cvpr/ChenFU14}
Liang{-}Chieh Chen, Sanja Fidler, and Raquel Urtasun.
\newblock Beat the mturkers: Automatic image labeling from weak 3d supervision.
\newblock In \emph{2014 {IEEE} Conference on Computer Vision and Pattern
  Recognition, {CVPR} 2014, Columbus, OH, USA, June 23-28, 2014}, pages
  3198--3205, 2014.
\newblock \doi{10.1109/CVPR.2014.409}.
\newblock URL \url{https://doi.org/10.1109/CVPR.2014.409}.

\bibitem[Cohn et~al.(1994)Cohn, Ghahramani, and
  Jordan]{DBLP:conf/nips/CohnGJ94}
David~A. Cohn, Zoubin Ghahramani, and Michael~I. Jordan.
\newblock Active learning with statistical models.
\newblock In \emph{Advances in Neural Information Processing Systems 7, {[NIPS}
  Conference, Denver, Colorado, USA, 1994]}, pages 705--712, 1994.
\newblock URL
  \url{http://papers.nips.cc/paper/1011-active-learning-with-statistical-models}.

\bibitem[Cordts et~al.(2016)Cordts, Omran, Ramos, Rehfeld, Enzweiler, Benenson,
  Franke, Roth, and Schiele]{DBLP:conf/cvpr/CordtsORREBFRS16}
Marius Cordts, Mohamed Omran, Sebastian Ramos, Timo Rehfeld, Markus Enzweiler,
  Rodrigo Benenson, Uwe Franke, Stefan Roth, and Bernt Schiele.
\newblock The cityscapes dataset for semantic urban scene understanding.
\newblock In \emph{2016 {IEEE} Conference on Computer Vision and Pattern
  Recognition, {CVPR} 2016, Las Vegas, NV, USA, June 27-30, 2016}, pages
  3213--3223, 2016.
\newblock \doi{10.1109/CVPR.2016.350}.
\newblock URL \url{https://doi.org/10.1109/CVPR.2016.350}.

\bibitem[Dagan and Engelson(1995)]{DBLP:conf/icml/DaganE95}
Ido Dagan and Sean~P. Engelson.
\newblock Committee-based sampling for training probabilistic classifiers.
\newblock In \emph{Machine Learning, Proceedings of the Twelfth International
  Conference on Machine Learning, Tahoe City, California, USA, July 9-12,
  1995}, pages 150--157, 1995.
\newblock URL
  \url{http://citeseerx.ist.psu.edu/viewdoc/summary?doi=10.1.1.30.6148}.

\bibitem[Deng et~al.(2009)Deng, Dong, Socher, Li, Li, and
  Li]{DBLP:conf/cvpr/DengDSLL009}
Jia Deng, Wei Dong, Richard Socher, Li{-}Jia Li, Kai Li, and Fei{-}Fei Li.
\newblock Imagenet: {A} large-scale hierarchical image database.
\newblock In \emph{2009 {IEEE} Computer Society Conference on Computer Vision
  and Pattern Recognition {(CVPR} 2009), 20-25 June 2009, Miami, Florida,
  {USA}}, pages 248--255, 2009.
\newblock \doi{10.1109/CVPRW.2009.5206848}.
\newblock URL \url{https://doi.org/10.1109/CVPRW.2009.5206848}.

\bibitem[Fathi et~al.(2011)Fathi, Balcan, Ren, and
  Rehg]{DBLP:conf/bmvc/FathiBRR11}
Alireza Fathi, Maria{-}Florina Balcan, Xiaofeng Ren, and James~M. Rehg.
\newblock Combining self training and active learning for video segmentation.
\newblock In \emph{British Machine Vision Conference, {BMVC} 2011, Dundee, UK,
  August 29 - September 2, 2011. Proceedings}, pages 1--11, 2011.
\newblock \doi{10.5244/C.25.78}.
\newblock URL \url{https://doi.org/10.5244/C.25.78}.

\bibitem[Feng et~al.(2016)Feng, Price, Cohen, and
  Chang]{DBLP:conf/cvpr/FengPCC16}
Jie Feng, Brian~L. Price, Scott Cohen, and Shih{-}Fu Chang.
\newblock Interactive segmentation on {RGBD} images via cue selection.
\newblock In \emph{2016 {IEEE} Conference on Computer Vision and Pattern
  Recognition, {CVPR} 2016, Las Vegas, NV, USA, June 27-30, 2016}, pages
  156--164, 2016.
\newblock \doi{10.1109/CVPR.2016.24}.
\newblock URL \url{https://doi.org/10.1109/CVPR.2016.24}.

\bibitem[Freund et~al.(1992)Freund, Seung, Shamir, and
  Tishby]{DBLP:conf/nips/FreundSST92}
Yoav Freund, H.~Sebastian Seung, Eli Shamir, and Naftali Tishby.
\newblock Information, prediction, and query by committee.
\newblock In \emph{Advances in Neural Information Processing Systems 5, {[NIPS}
  Conference, Denver, Colorado, USA, November 30 - December 3, 1992]}, pages
  483--490, 1992.
\newblock URL
  \url{http://papers.nips.cc/paper/622-information-prediction-and-query-by-committee}.

\bibitem[Gal and Ghahramani(2016)]{DBLP:conf/icml/GalG16}
Yarin Gal and Zoubin Ghahramani.
\newblock Dropout as a bayesian approximation: Representing model uncertainty
  in deep learning.
\newblock In \emph{Proceedings of the 33nd International Conference on Machine
  Learning, {ICML} 2016, New York City, NY, USA, June 19-24, 2016}, pages
  1050--1059, 2016.
\newblock URL \url{http://jmlr.org/proceedings/papers/v48/gal16.html}.

\bibitem[Gal et~al.(2017)Gal, Islam, and Ghahramani]{DBLP:conf/icml/GalIG17}
Yarin Gal, Riashat Islam, and Zoubin Ghahramani.
\newblock Deep bayesian active learning with image data.
\newblock In \emph{Proceedings of the 34th International Conference on Machine
  Learning, {ICML} 2017, Sydney, NSW, Australia, 6-11 August 2017}, pages
  1183--1192, 2017.
\newblock URL \url{http://proceedings.mlr.press/v70/gal17a.html}.

\bibitem[Girshick et~al.(2014)Girshick, Donahue, Darrell, and
  Malik]{DBLP:conf/cvpr/GirshickDDM14}
Ross~B. Girshick, Jeff Donahue, Trevor Darrell, and Jitendra Malik.
\newblock Rich feature hierarchies for accurate object detection and semantic
  segmentation.
\newblock In \emph{2014 {IEEE} Conference on Computer Vision and Pattern
  Recognition, {CVPR} 2014, Columbus, OH, USA, June 23-28, 2014}, pages
  580--587, 2014.
\newblock \doi{10.1109/CVPR.2014.81}.
\newblock URL \url{https://doi.org/10.1109/CVPR.2014.81}.

\bibitem[Gorriz et~al.(2017)Gorriz, Carlier, Faure, and {Gir{\'{o}} i
  Nieto}]{DBLP:journals/corr/abs-1711-09168}
Marc Gorriz, Axel Carlier, Emmanuel Faure, and Xavier {Gir{\'{o}} i Nieto}.
\newblock Cost-effective active learning for melanoma segmentation.
\newblock \emph{CoRR}, abs/1711.09168, 2017.
\newblock URL \url{http://arxiv.org/abs/1711.09168}.

\bibitem[Hong et~al.(2017)Hong, Yeo, Kwak, Lee, and
  Han]{DBLP:conf/cvpr/HongYKLH17}
Seunghoon Hong, Donghun Yeo, Suha Kwak, Honglak Lee, and Bohyung Han.
\newblock Weakly supervised semantic segmentation using web-crawled videos.
\newblock In \emph{2017 {IEEE} Conference on Computer Vision and Pattern
  Recognition, {CVPR} 2017, Honolulu, HI, USA, July 21-26, 2017}, pages
  2224--2232, 2017.
\newblock \doi{10.1109/CVPR.2017.239}.
\newblock URL \url{https://doi.org/10.1109/CVPR.2017.239}.

\bibitem[Howard et~al.(2017)Howard, Zhu, Chen, Kalenichenko, Wang, Weyand,
  Andreetto, and Adam]{DBLP:journals/corr/HowardZCKWWAA17}
Andrew~G. Howard, Menglong Zhu, Bo~Chen, Dmitry Kalenichenko, Weijun Wang,
  Tobias Weyand, Marco Andreetto, and Hartwig Adam.
\newblock Mobilenets: Efficient convolutional neural networks for mobile vision
  applications.
\newblock \emph{CoRR}, abs/1704.04861, 2017.
\newblock URL \url{http://arxiv.org/abs/1704.04861}.

\bibitem[Huh et~al.(2016)Huh, Agrawal, and Efros]{DBLP:journals/corr/HuhAE16}
Mi{-}Young Huh, Pulkit Agrawal, and Alexei~A. Efros.
\newblock What makes imagenet good for transfer learning?
\newblock \emph{CoRR}, abs/1608.08614, 2016.
\newblock URL \url{http://arxiv.org/abs/1608.08614}.

\bibitem[Hung et~al.(2018)Hung, Tsai, Liou, Lin, and
  Yang]{DBLP:journals/corr/abs-1802-07934}
Wei{-}Chih Hung, Yi{-}Hsuan Tsai, Yan{-}Ting Liou, Yen{-}Yu Lin, and
  Ming{-}Hsuan Yang.
\newblock Adversarial learning for semi-supervised semantic segmentation.
\newblock \emph{CoRR}, abs/1802.07934, 2018.
\newblock URL \url{http://arxiv.org/abs/1802.07934}.

\bibitem[Jain and Grauman(2016)]{DBLP:conf/cvpr/JainG16}
Suyog~Dutt Jain and Kristen Grauman.
\newblock Active image segmentation propagation.
\newblock In \emph{2016 {IEEE} Conference on Computer Vision and Pattern
  Recognition, {CVPR} 2016, Las Vegas, NV, USA, June 27-30, 2016}, pages
  2864--2873, 2016.
\newblock \doi{10.1109/CVPR.2016.313}.
\newblock URL \url{https://doi.org/10.1109/CVPR.2016.313}.

\bibitem[Khoreva et~al.(2017)Khoreva, Benenson, Hosang, Hein, and
  Schiele]{DBLP:conf/cvpr/KhorevaBH0S17}
Anna Khoreva, Rodrigo Benenson, Jan~Hendrik Hosang, Matthias Hein, and Bernt
  Schiele.
\newblock Simple does it: Weakly supervised instance and semantic segmentation.
\newblock In \emph{2017 {IEEE} Conference on Computer Vision and Pattern
  Recognition, {CVPR} 2017, Honolulu, HI, USA, July 21-26, 2017}, pages
  1665--1674, 2017.
\newblock \doi{10.1109/CVPR.2017.181}.
\newblock URL \url{https://doi.org/10.1109/CVPR.2017.181}.

\bibitem[Kirillov et~al.(2017)Kirillov, Levinkov, Andres, Savchynskyy, and
  Rother]{DBLP:conf/cvpr/KirillovLASR17}
Alexander Kirillov, Evgeny Levinkov, Bjoern Andres, Bogdan Savchynskyy, and
  Carsten Rother.
\newblock Instancecut: From edges to instances with multicut.
\newblock In \emph{2017 {IEEE} Conference on Computer Vision and Pattern
  Recognition, {CVPR} 2017, Honolulu, HI, USA, July 21-26, 2017}, pages
  7322--7331, 2017.
\newblock \doi{10.1109/CVPR.2017.774}.
\newblock URL \url{https://doi.org/10.1109/CVPR.2017.774}.

\bibitem[Kolesnikov and Lampert(2016)]{DBLP:conf/eccv/KolesnikovL16}
Alexander Kolesnikov and Christoph~H. Lampert.
\newblock Seed, expand and constrain: Three principles for weakly-supervised
  image segmentation.
\newblock In \emph{Computer Vision - {ECCV} 2016 - 14th European Conference,
  Amsterdam, The Netherlands, October 11-14, 2016, Proceedings, Part {IV}},
  pages 695--711, 2016.
\newblock \doi{10.1007/978-3-319-46493-0_42}.
\newblock URL \url{https://doi.org/10.1007/978-3-319-46493-0_42}.

\bibitem[Konyushkova et~al.(2015)Konyushkova, Sznitman, and
  Fua]{DBLP:conf/iccv/KonyushkovaSF15}
Ksenia Konyushkova, Raphael Sznitman, and Pascal Fua.
\newblock Introducing geometry in active learning for image segmentation.
\newblock In \emph{2015 {IEEE} International Conference on Computer Vision,
  {ICCV} 2015, Santiago, Chile, December 7-13, 2015}, pages 2974--2982, 2015.
\newblock \doi{10.1109/ICCV.2015.340}.
\newblock URL \url{https://doi.org/10.1109/ICCV.2015.340}.

\bibitem[Kovashka et~al.(2016)Kovashka, Russakovsky, Fei{-}Fei, and
  Grauman]{DBLP:journals/ftcgv/KovashkaRFG16}
Adriana Kovashka, Olga Russakovsky, Li~Fei{-}Fei, and Kristen Grauman.
\newblock Crowdsourcing in computer vision.
\newblock \emph{Foundations and Trends in Computer Graphics and Vision},
  10\penalty0 (3):\penalty0 177--243, 2016.
\newblock \doi{10.1561/0600000071}.
\newblock URL \url{https://doi.org/10.1561/0600000071}.

\bibitem[Krishnamurthy et~al.(2017)Krishnamurthy, Agarwal, Huang, III, and
  Langford]{DBLP:conf/icml/KrishnamurthyAH17}
Akshay Krishnamurthy, Alekh Agarwal, Tzu{-}Kuo Huang, Hal~Daum{\'{e}} III, and
  John Langford.
\newblock Active learning for cost-sensitive classification.
\newblock In \emph{Proceedings of the 34th International Conference on Machine
  Learning, {ICML} 2017, Sydney, NSW, Australia, 6-11 August 2017}, pages
  1915--1924, 2017.
\newblock URL \url{http://proceedings.mlr.press/v70/krishnamurthy17a.html}.

\bibitem[Lewis(1995)]{DBLP:journals/sigir/Lewis95a}
David~D. Lewis.
\newblock A sequential algorithm for training text classifiers: Corrigendum and
  additional data.
\newblock \emph{{SIGIR} Forum}, 29\penalty0 (2):\penalty0 13--19, 1995.
\newblock \doi{10.1145/219587.219592}.
\newblock URL \url{http://doi.acm.org/10.1145/219587.219592}.

\bibitem[Long et~al.(2015)Long, Shelhamer, and
  Darrell]{DBLP:conf/cvpr/LongSD15}
Jonathan Long, Evan Shelhamer, and Trevor Darrell.
\newblock Fully convolutional networks for semantic segmentation.
\newblock In \emph{{IEEE} Conference on Computer Vision and Pattern
  Recognition, {CVPR} 2015, Boston, MA, USA, June 7-12, 2015}, pages
  3431--3440, 2015.
\newblock \doi{10.1109/CVPR.2015.7298965}.
\newblock URL \url{https://doi.org/10.1109/CVPR.2015.7298965}.

\bibitem[Lu et~al.(2016)Lu, Bai, Shapiro, and Wang]{DBLP:conf/cvpr/LuBSW16}
Yao Lu, Xue Bai, Linda~G. Shapiro, and Jue Wang.
\newblock Coherent parametric contours for interactive video object
  segmentation.
\newblock In \emph{2016 {IEEE} Conference on Computer Vision and Pattern
  Recognition, {CVPR} 2016, Las Vegas, NV, USA, June 27-30, 2016}, pages
  642--650, 2016.
\newblock \doi{10.1109/CVPR.2016.76}.
\newblock URL \url{https://doi.org/10.1109/CVPR.2016.76}.

\bibitem[Lu et~al.(2017)Lu, Fu, Xiang, Han, Wang, and
  Gao]{DBLP:journals/pami/LuFXHWG17}
Zhiwu Lu, Zhenyong Fu, Tao Xiang, Peng Han, Liwei Wang, and Xin Gao.
\newblock Learning from weak and noisy labels for semantic segmentation.
\newblock \emph{{IEEE} Trans. Pattern Anal. Mach. Intell.}, 39\penalty0
  (3):\penalty0 486--500, 2017.
\newblock \doi{10.1109/TPAMI.2016.2552172}.
\newblock URL \url{https://doi.org/10.1109/TPAMI.2016.2552172}.

\bibitem[Mosinska{-}Domanska et~al.(2016)Mosinska{-}Domanska, Sznitman,
  Glowacki, and Fua]{DBLP:conf/cvpr/Mosinska-Domanska16}
Agata Mosinska{-}Domanska, Raphael Sznitman, Przemyslaw Glowacki, and Pascal
  Fua.
\newblock Active learning for delineation of curvilinear structures.
\newblock In \emph{2016 {IEEE} Conference on Computer Vision and Pattern
  Recognition, {CVPR} 2016, Las Vegas, NV, USA, June 27-30, 2016}, pages
  5231--5239, 2016.
\newblock \doi{10.1109/CVPR.2016.565}.
\newblock URL \url{https://doi.org/10.1109/CVPR.2016.565}.

\bibitem[Movshovitz{-}Attias et~al.(2016)Movshovitz{-}Attias, Kanade, and
  Sheikh]{DBLP:conf/eccv/Movshovitz-Attias16}
Yair Movshovitz{-}Attias, Takeo Kanade, and Yaser Sheikh.
\newblock How useful is photo-realistic rendering for visual learning?
\newblock In \emph{Computer Vision - {ECCV} 2016 Workshops - Amsterdam, The
  Netherlands, October 8-10 and 15-16, 2016, Proceedings, Part {III}}, pages
  202--217, 2016.
\newblock \doi{10.1007/978-3-319-49409-8_18}.
\newblock URL \url{https://doi.org/10.1007/978-3-319-49409-8_18}.

\bibitem[Neuhold et~al.(2017)Neuhold, Ollmann, Bul{\`{o}}, and
  Kontschieder]{DBLP:conf/iccv/NeuholdOBK17}
Gerhard Neuhold, Tobias Ollmann, Samuel~Rota Bul{\`{o}}, and Peter
  Kontschieder.
\newblock The mapillary vistas dataset for semantic understanding of street
  scenes.
\newblock In \emph{{IEEE} International Conference on Computer Vision, {ICCV}
  2017, Venice, Italy, October 22-29, 2017}, pages 5000--5009, 2017.
\newblock \doi{10.1109/ICCV.2017.534}.
\newblock URL \url{https://doi.org/10.1109/ICCV.2017.534}.

\bibitem[Oh et~al.(2017)Oh, Benenson, Khoreva, Akata, Fritz, and
  Schiele]{DBLP:conf/cvpr/OhBKAFS17}
Seong~Joon Oh, Rodrigo Benenson, Anna Khoreva, Zeynep Akata, Mario Fritz, and
  Bernt Schiele.
\newblock Exploiting saliency for object segmentation from image level labels.
\newblock In \emph{2017 {IEEE} Conference on Computer Vision and Pattern
  Recognition, {CVPR} 2017, Honolulu, HI, USA, July 21-26, 2017}, pages
  5038--5047, 2017.
\newblock \doi{10.1109/CVPR.2017.535}.
\newblock URL \url{https://doi.org/10.1109/CVPR.2017.535}.

\bibitem[Oquab et~al.(2014)Oquab, Bottou, Laptev, and
  Sivic]{DBLP:conf/cvpr/OquabBLS14}
Maxime Oquab, L{\'{e}}on Bottou, Ivan Laptev, and Josef Sivic.
\newblock Learning and transferring mid-level image representations using
  convolutional neural networks.
\newblock In \emph{2014 {IEEE} Conference on Computer Vision and Pattern
  Recognition, {CVPR} 2014, Columbus, OH, USA, June 23-28, 2014}, pages
  1717--1724, 2014.
\newblock \doi{10.1109/CVPR.2014.222}.
\newblock URL \url{https://doi.org/10.1109/CVPR.2014.222}.

\bibitem[Papandreou et~al.(2015)Papandreou, Chen, Murphy, and
  Yuille]{DBLP:conf/iccv/PapandreouCMY15}
George Papandreou, Liang{-}Chieh Chen, Kevin~P. Murphy, and Alan~L. Yuille.
\newblock Weakly-and semi-supervised learning of a deep convolutional network
  for semantic image segmentation.
\newblock In \emph{2015 {IEEE} International Conference on Computer Vision,
  {ICCV} 2015, Santiago, Chile, December 7-13, 2015}, pages 1742--1750, 2015.
\newblock \doi{10.1109/ICCV.2015.203}.
\newblock URL \url{https://doi.org/10.1109/ICCV.2015.203}.

\bibitem[Richter et~al.(2016)Richter, Vineet, Roth, and
  Koltun]{DBLP:conf/eccv/RichterVRK16}
Stephan~R. Richter, Vibhav Vineet, Stefan Roth, and Vladlen Koltun.
\newblock Playing for data: Ground truth from computer games.
\newblock In \emph{Computer Vision - {ECCV} 2016 - 14th European Conference,
  Amsterdam, The Netherlands, October 11-14, 2016, Proceedings, Part {II}},
  pages 102--118, 2016.
\newblock \doi{10.1007/978-3-319-46475-6_7}.
\newblock URL \url{https://doi.org/10.1007/978-3-319-46475-6_7}.

\bibitem[Russell et~al.(2008)Russell, Torralba, Murphy, and
  Freeman]{DBLP:journals/ijcv/RussellTMF08}
Bryan~C. Russell, Antonio Torralba, Kevin~P. Murphy, and William~T. Freeman.
\newblock Labelme: {A} database and web-based tool for image annotation.
\newblock \emph{International Journal of Computer Vision}, 77\penalty0
  (1-3):\penalty0 157--173, 2008.
\newblock \doi{10.1007/s11263-007-0090-8}.
\newblock URL \url{https://doi.org/10.1007/s11263-007-0090-8}.

\bibitem[Saleh et~al.(2016)Saleh, Akbarian, Salzmann, Petersson, Gould, and
  Alvarez]{DBLP:conf/eccv/SalehASPGA16}
Fatemehsadat Saleh, Mohammad Sadegh~Ali Akbarian, Mathieu Salzmann, Lars
  Petersson, Stephen Gould, and Jose~M. Alvarez.
\newblock Built-in foreground/background prior for weakly-supervised semantic
  segmentation.
\newblock In \emph{Computer Vision - {ECCV} 2016 - 14th European Conference,
  Amsterdam, The Netherlands, October 11-14, 2016, Proceedings, Part {VIII}},
  pages 413--432, 2016.
\newblock \doi{10.1007/978-3-319-46484-8_25}.
\newblock URL \url{https://doi.org/10.1007/978-3-319-46484-8_25}.

\bibitem[Saleh et~al.(2017)Saleh, Akbarian, Salzmann, Petersson, and
  Alvarez]{DBLP:conf/iccv/SalehASPA17}
Fatemehsadat Saleh, Mohammad Sadegh~Ali Akbarian, Mathieu Salzmann, Lars
  Petersson, and Jose~M. Alvarez.
\newblock Bringing background into the foreground: Making all classes equal in
  weakly-supervised video semantic segmentation.
\newblock In \emph{{IEEE} International Conference on Computer Vision, {ICCV}
  2017, Venice, Italy, October 22-29, 2017}, pages 2125--2135, 2017.
\newblock \doi{10.1109/ICCV.2017.232}.
\newblock URL \url{https://doi.org/10.1109/ICCV.2017.232}.

\bibitem[Sener and Savarese(2017)]{DBLP:journals/corr/abs-1708-00489}
Ozan Sener and Silvio Savarese.
\newblock A geometric approach to active learning for convolutional neural
  networks.
\newblock \emph{CoRR}, abs/1708.00489, 2017.
\newblock URL \url{http://arxiv.org/abs/1708.00489}.

\bibitem[Sener and Savarese(2018)]{sener2018active}
Ozan Sener and Silvio Savarese.
\newblock Active learning for convolutional neural networks: A core-set
  approach.
\newblock In \emph{International Conference on Learning Representations}, 2018.
\newblock URL \url{https://openreview.net/forum?id=H1aIuk-RW}.

\bibitem[Settles(2009)]{settles2009active}
Burr Settles.
\newblock Active learning literature survey.
\newblock Computer Sciences Technical Report 1648, University of
  Wisconsin--Madison, 2009.
\newblock URL
  \url{http://axon.cs.byu.edu/~martinez/classes/778/Papers/settles.activelearning.pdf}.

\bibitem[Settles et~al.(2008)Settles, Craven, and
  Friedl]{Settles_activelearning}
Burr Settles, Mark Craven, and Lewis Friedl.
\newblock Active learning with real annotation costs.
\newblock In \emph{In Proceedings of the NIPS Workshop on Cost-Sensitive
  Learning}, pages 1--10, 2008.
\newblock URL \url{http://burrsettles.com/pub/settles.nips08ws.pdf}.

\bibitem[Shafaei et~al.(2016)Shafaei, Little, and
  Schmidt]{DBLP:conf/bmvc/ShafaeiLS16}
Alireza Shafaei, James~J. Little, and Mark Schmidt.
\newblock Play and learn: Using video games to train computer vision models.
\newblock In \emph{Proceedings of the British Machine Vision Conference 2016,
  {BMVC} 2016, York, UK, September 19-22, 2016}, 2016.
\newblock URL \url{http://www.bmva.org/bmvc/2016/papers/paper026/index.html}.

\bibitem[Shannon(2001)]{DBLP:journals/sigmobile/Shannon01}
Claude~E. Shannon.
\newblock A mathematical theory of communication.
\newblock \emph{Mobile Computing and Communications Review}, 5\penalty0
  (1):\penalty0 3--55, 2001.
\newblock \doi{10.1145/584091.584093}.
\newblock URL \url{http://doi.acm.org/10.1145/584091.584093}.

\bibitem[Shimoda and Yanai(2016)]{DBLP:conf/eccv/ShimodaY16}
Wataru Shimoda and Keiji Yanai.
\newblock Distinct class-specific saliency maps for weakly supervised semantic
  segmentation.
\newblock In \emph{Computer Vision - {ECCV} 2016 - 14th European Conference,
  Amsterdam, The Netherlands, October 11-14, 2016, Proceedings, Part {IV}},
  pages 218--234, 2016.
\newblock \doi{10.1007/978-3-319-46493-0_14}.
\newblock URL \url{https://doi.org/10.1007/978-3-319-46493-0_14}.

\bibitem[Souly et~al.(2017)Souly, Spampinato, and Shah]{8237868}
Nasim Souly, Concetto Spampinato, and Mubarak Shah.
\newblock Semi supervised semantic segmentation using generative adversarial
  network.
\newblock In \emph{{IEEE} International Conference on Computer Vision, {ICCV}
  2017, Venice, Italy, October 22-29, 2017}, pages 5689--5697, 2017.
\newblock \doi{10.1109/ICCV.2017.606}.
\newblock URL \url{https://doi.org/10.1109/ICCV.2017.606}.

\bibitem[Sun et~al.(2017)Sun, Shrivastava, Singh, and
  Gupta]{DBLP:conf/iccv/SunSSG17}
Chen Sun, Abhinav Shrivastava, Saurabh Singh, and Abhinav Gupta.
\newblock Revisiting unreasonable effectiveness of data in deep learning era.
\newblock In \emph{{IEEE} International Conference on Computer Vision, {ICCV}
  2017, Venice, Italy, October 22-29, 2017}, pages 843--852, 2017.
\newblock \doi{10.1109/ICCV.2017.97}.
\newblock URL \url{https://doi.org/10.1109/ICCV.2017.97}.

\bibitem[Tong and Koller(2000)]{DBLP:conf/nips/TongK00}
Simon Tong and Daphne Koller.
\newblock Active learning for parameter estimation in bayesian networks.
\newblock In \emph{Advances in Neural Information Processing Systems 13, Papers
  from Neural Information Processing Systems {(NIPS)} 2000, Denver, CO, {USA}},
  pages 647--653, 2000.
\newblock URL
  \url{http://papers.nips.cc/paper/1795-active-learning-for-parameter-estimation-in-bayesian-networks}.

\bibitem[Tong and Koller(2001)]{DBLP:journals/jmlr/TongK01}
Simon Tong and Daphne Koller.
\newblock Support vector machine active learning with applications to text
  classification.
\newblock \emph{Journal of Machine Learning Research}, 2:\penalty0 45--66,
  2001.
\newblock URL
  \url{http://www.ai.mit.edu/projects/jmlr/papers/volume2/tong01a/abstract.html}.

\bibitem[Vezhnevets et~al.(2012)Vezhnevets, Buhmann, and
  Ferrari]{DBLP:conf/cvpr/VezhnevetsBF12}
Alexander Vezhnevets, Joachim~M. Buhmann, and Vittorio Ferrari.
\newblock Active learning for semantic segmentation with expected change.
\newblock In \emph{2012 {IEEE} Conference on Computer Vision and Pattern
  Recognition, Providence, RI, USA, June 16-21, 2012}, pages 3162--3169, 2012.
\newblock \doi{10.1109/CVPR.2012.6248050}.
\newblock URL \url{https://doi.org/10.1109/CVPR.2012.6248050}.

\bibitem[Vijayanarasimhan and
  Grauman(2009)]{DBLP:conf/cvpr/VijayanarasimhanG09}
Sudheendra Vijayanarasimhan and Kristen Grauman.
\newblock What's it going to cost you?: Predicting effort vs. informativeness
  for multi-label image annotations.
\newblock In \emph{2009 {IEEE} Computer Society Conference on Computer Vision
  and Pattern Recognition {(CVPR} 2009), 20-25 June 2009, Miami, Florida,
  {USA}}, pages 2262--2269, 2009.
\newblock \doi{10.1109/CVPRW.2009.5206705}.
\newblock URL \url{https://doi.org/10.1109/CVPRW.2009.5206705}.

\bibitem[Wang et~al.(2017)Wang, Zhang, Li, Zhang, and
  Lin]{DBLP:journals/tcsv/WangZLZL17}
Keze Wang, Dongyu Zhang, Ya~Li, Ruimao Zhang, and Liang Lin.
\newblock Cost-effective active learning for deep image classification.
\newblock \emph{{IEEE} Trans. Circuits Syst. Video Techn.}, 27\penalty0
  (12):\penalty0 2591--2600, 2017.
\newblock \doi{10.1109/TCSVT.2016.2589879}.
\newblock URL \url{https://doi.org/10.1109/TCSVT.2016.2589879}.

\bibitem[Xie et~al.(2016)Xie, Kiefel, Sun, and Geiger]{DBLP:conf/cvpr/XieKSG16}
Jun Xie, Martin Kiefel, Ming{-}Ting Sun, and Andreas Geiger.
\newblock Semantic instance annotation of street scenes by 3d to 2d label
  transfer.
\newblock In \emph{2016 {IEEE} Conference on Computer Vision and Pattern
  Recognition, {CVPR} 2016, Las Vegas, NV, USA, June 27-30, 2016}, pages
  3688--3697, 2016.
\newblock \doi{10.1109/CVPR.2016.401}.
\newblock URL \url{https://doi.org/10.1109/CVPR.2016.401}.

\bibitem[Xu et~al.(2016)Xu, Price, Cohen, Yang, and
  Huang]{DBLP:conf/cvpr/XuPCYH16}
Ning Xu, Brian~L. Price, Scott Cohen, Jimei Yang, and Thomas~S. Huang.
\newblock Deep interactive object selection.
\newblock In \emph{2016 {IEEE} Conference on Computer Vision and Pattern
  Recognition, {CVPR} 2016, Las Vegas, NV, USA, June 27-30, 2016}, pages
  373--381, 2016.
\newblock \doi{10.1109/CVPR.2016.47}.
\newblock URL \url{https://doi.org/10.1109/CVPR.2016.47}.

\bibitem[Yang et~al.(2017)Yang, Zhang, Chen, Zhang, and
  Chen]{DBLP:conf/miccai/YangZCZC17}
Lin Yang, Yizhe Zhang, Jianxu Chen, Siyuan Zhang, and Danny~Z. Chen.
\newblock Suggestive annotation: {A} deep active learning framework for
  biomedical image segmentation.
\newblock In \emph{Medical Image Computing and Computer Assisted Intervention -
  {MICCAI} 2017 - 20th International Conference, Quebec City, QC, Canada,
  September 11-13, 2017, Proceedings, Part {III}}, pages 399--407, 2017.
\newblock \doi{10.1007/978-3-319-66179-7_46}.
\newblock URL \url{https://doi.org/10.1007/978-3-319-66179-7_46}.

\bibitem[Zhang et~al.(2015)Zhang, Zeng, Wang, and
  Xue]{DBLP:conf/cvpr/ZhangZWX15}
Wei Zhang, Sheng Zeng, Dequan Wang, and Xiangyang Xue.
\newblock Weakly supervised semantic segmentation for social images.
\newblock In \emph{{IEEE} Conference on Computer Vision and Pattern
  Recognition, {CVPR} 2015, Boston, MA, USA, June 7-12, 2015}, pages
  2718--2726, 2015.
\newblock \doi{10.1109/CVPR.2015.7298888}.
\newblock URL \url{https://doi.org/10.1109/CVPR.2015.7298888}.

\bibitem[Zhang et~al.(2017)Zhang, Song, Yumer, Savva, Lee, Jin, and
  Funkhouser]{DBLP:conf/cvpr/ZhangSYSLJF17}
Yinda Zhang, Shuran Song, Ersin Yumer, Manolis Savva, Joon{-}Young Lee, Hailin
  Jin, and Thomas~A. Funkhouser.
\newblock Physically-based rendering for indoor scene understanding using
  convolutional neural networks.
\newblock In \emph{2017 {IEEE} Conference on Computer Vision and Pattern
  Recognition, {CVPR} 2017, Honolulu, HI, USA, July 21-26, 2017}, pages
  5057--5065, 2017.
\newblock \doi{10.1109/CVPR.2017.537}.
\newblock URL \url{https://doi.org/10.1109/CVPR.2017.537}.

\bibitem[Zhou et~al.(2017)Zhou, Zhao, Puig, Fidler, Barriuso, and
  Torralba]{DBLP:conf/cvpr/ZhouZPFB017}
Bolei Zhou, Hang Zhao, Xavier Puig, Sanja Fidler, Adela Barriuso, and Antonio
  Torralba.
\newblock Scene parsing through {ADE20K} dataset.
\newblock In \emph{2017 {IEEE} Conference on Computer Vision and Pattern
  Recognition, {CVPR} 2017, Honolulu, HI, USA, July 21-26, 2017}, pages
  5122--5130, 2017.
\newblock \doi{10.1109/CVPR.2017.544}.
\newblock URL \url{https://doi.org/10.1109/CVPR.2017.544}.

\end{thebibliography}

\clearpage

\appendix

\section{Supplementary Material \newline CEREALS -- Cost-Effective REgion-based Active Learning for Semantic Segmentation}

\subsection{Implementation Details}
\label{appendix:impl_details}

Instead of cropping the annotated regions out of the images, while taking into account their receptive field in input space, we instead mask out all currently unlabeled data in output space, making sure that no loss is computed on unlabeled data when learning the \textit{semantic segmentation model} nor when learning the \textit{cost model}. We then perform an image-based training, from unprocessed input images to spatial label maps. However, our practical implementation of \textit{CEREALS}, which will be made publicly available is supporting both options. For training the utilized models we use Adam as our optimizer with learning rate, alpha and beta set to $0.0001$, $0.99$ and $0.999$ respectively. Furthermore, we claim convergence whenever a model hasn't improved regarding the application loss for at least $10$ epochs. We train with the mini-batch size set to $1$, such that a gradient step is always being applied w.r.t. one full resolution image of Cityscapes.

\paragraph{Semantic Segmentation Model}
We do not train the employed model in stages, but directly optimize for \textit{FCN8s}. Regarding the training performed on the full training set of Cityscapes, we report a mean intersection over union (mIoU) of $0.605$ which is, as all other results, computed on the full validation dataset of Cityscapes. Note, that the original model achieves a mIoU of $0.65$ and that we are able to reproduce this result when the width multiplier is set to $1.0$, despite all other changes. Though we utilized this particular model, \textit{CEREALS} can use any model producing semantic segmentation masks as long as it provides probability distributions regarding it's posterior outcome. The \textit{cost model} however, would need to be adapted or made independent of the \textit{semantic segmentation model} in such a case. 

\paragraph{Cost Model}
The only change we made to the original model's architecture is to replace it's softmax activation with a linear activation layer. We trained the model towards minimizing the mean squared error of predicted and ground truth clicks. Since we observed some pixels to have unrealistically many clicks in the ground truth data, we clipped the values to be in $[0,10]$ range allowing a maximum of 10 ground truth clicks per pixel. As the \textit{semantic segmentation model}, the \textit{cost model} doesn't have any upsampling layer at the end, in order to allow for faster trainings. We instead downscale the provided click data by a factor of $8$. 

\clearpage
\subsection{Results}
\label{appendix:results}
The evaluation for all processed experiments performed on the modified \textit{FCN8s} architecture are reported in Table \ref{supp:table}.

\begin{table}[ht]
\resizebox{\textwidth}{!}{%
\bgroup
\def\arraystretch{1.25}
\begin{tabular}{|c|c|c|c|c|cccccccccc}
\hline
\multicolumn{1}{|l|}{}          & \textbf{2048x1024*} & \textbf{512x512} & \textbf{256x256} & \multicolumn{11}{c|}{\textbf{128x128}}                                                                                                                                                                                                                                                                                                                                        \\ \hline
\textbf{Eq. \ref{eq2}) H (p95)} & 38.66                 & 16.74            & 11.71            & 10.01 & \multicolumn{2}{c|}{\textbf{Eq.\ref{eq6})}}                           & \multicolumn{2}{c|}{\textbf{Eq.\ref{eq7})}}                           & \multicolumn{2}{c|}{\textbf{Eq.\ref{eq8} ($\alpha=0.5$)}}            & \multicolumn{2}{c|}{\textbf{Eq.\ref{eq8} ($\alpha=0.75$)}}           & \multicolumn{2}{c|}{\textbf{Eq.\ref{eq8} ($\alpha=0.9$)}}            \\ \cline{6-15} 
\textbf{Eq. \ref{eq3}) V (p95)} & 36.97                 & 19.86            & 14.85            & 11.59 & \multicolumn{1}{c|}{\textbf{GT}} & \multicolumn{1}{c|}{\textbf{Est.}} & \multicolumn{1}{c|}{\textbf{GT}} & \multicolumn{1}{c|}{\textbf{Est.}} & \multicolumn{1}{c|}{\textbf{GT}} & \multicolumn{1}{c|}{\textbf{Est.}} & \multicolumn{1}{c|}{\textbf{GT}} & \multicolumn{1}{c|}{\textbf{Est.}} & \multicolumn{1}{c|}{\textbf{GT}} & \multicolumn{1}{c|}{\textbf{Est.}} \\ \hline
\textbf{Eq. \ref{eq2}) H (c95)} & 43.17                & 33.55            & 34.05           & 33.76 & \multicolumn{1}{c|}{23.08}       & \multicolumn{1}{c|}{21.42}         & \multicolumn{1}{c|}{14.68}       & \multicolumn{1}{c|}{\textbf{17.07}}         & \multicolumn{1}{c|}{10.91}         & \multicolumn{1}{c|}{n/a}           & \multicolumn{1}{c|}{22.3}        & \multicolumn{1}{c|}{20.54}         & \multicolumn{1}{c|}{27.79}       & \multicolumn{1}{c|}{34.52}         \\
\textbf{Eq. \ref{eq3}) V (c95)} & 39.02                 & 29.58            & 28.11            & 24.81 & \multicolumn{1}{c|}{15.30}       & \multicolumn{1}{c|}{18.23}             & \multicolumn{1}{c|}{13.97}       & \multicolumn{1}{c|}{18.41}             & \multicolumn{1}{c|}{n/a}         & \multicolumn{1}{c|}{n/a}           & \multicolumn{1}{c|}{16.35}       & \multicolumn{1}{c|}{19.01}             & \multicolumn{1}{c|}{20.11}       & \multicolumn{1}{c|}{21.56}             \\ \hline
\textbf{Random (p95)}           & n/a                 & 33.61              & 35.29              & 28.57 & \multicolumn{1}{l}{}             & \multicolumn{1}{l}{}               & \multicolumn{1}{l}{}             & \multicolumn{1}{l}{}               & \multicolumn{1}{l}{}             & \multicolumn{1}{l}{}               & \multicolumn{1}{l}{}             & \multicolumn{1}{l}{}               & \multicolumn{1}{l}{}             & \multicolumn{1}{l}{}               \\
\textbf{Random (c95)}           & n/a                 & 44.57              & 38.62              & 29.86 & \multicolumn{1}{l}{}             & \multicolumn{1}{l}{}               & \multicolumn{1}{l}{}             & \multicolumn{1}{l}{}               & \multicolumn{1}{l}{}             & \multicolumn{1}{l}{}               & \multicolumn{1}{l}{}             & \multicolumn{1}{l}{}               & \multicolumn{1}{l}{}             & \multicolumn{1}{l}{}               \\ \cline{1-5}
\end{tabular}
\egroup}
\vspace{0.01cm}
\caption{All results indicate the amount of used annotated pixels (p95) or number of clicks (c95) relative to the amount of all the pixels or clicks within the Cityscapes training set for achieving a mIoU of 95\% as compared to the mIoU achieved when learning the employed \textit{FCN8s}-based model on all the data provided by Cityscapes. The results in this table are averaged  over five repetitions, as explained in the beginning of the result section in this work. (*Image-based acquisition)}
\label{supp:table}
\end{table}

\begin{figure}[htb!]
\centering
\begin{subfigure}{\includegraphics[width=0.91\textwidth]{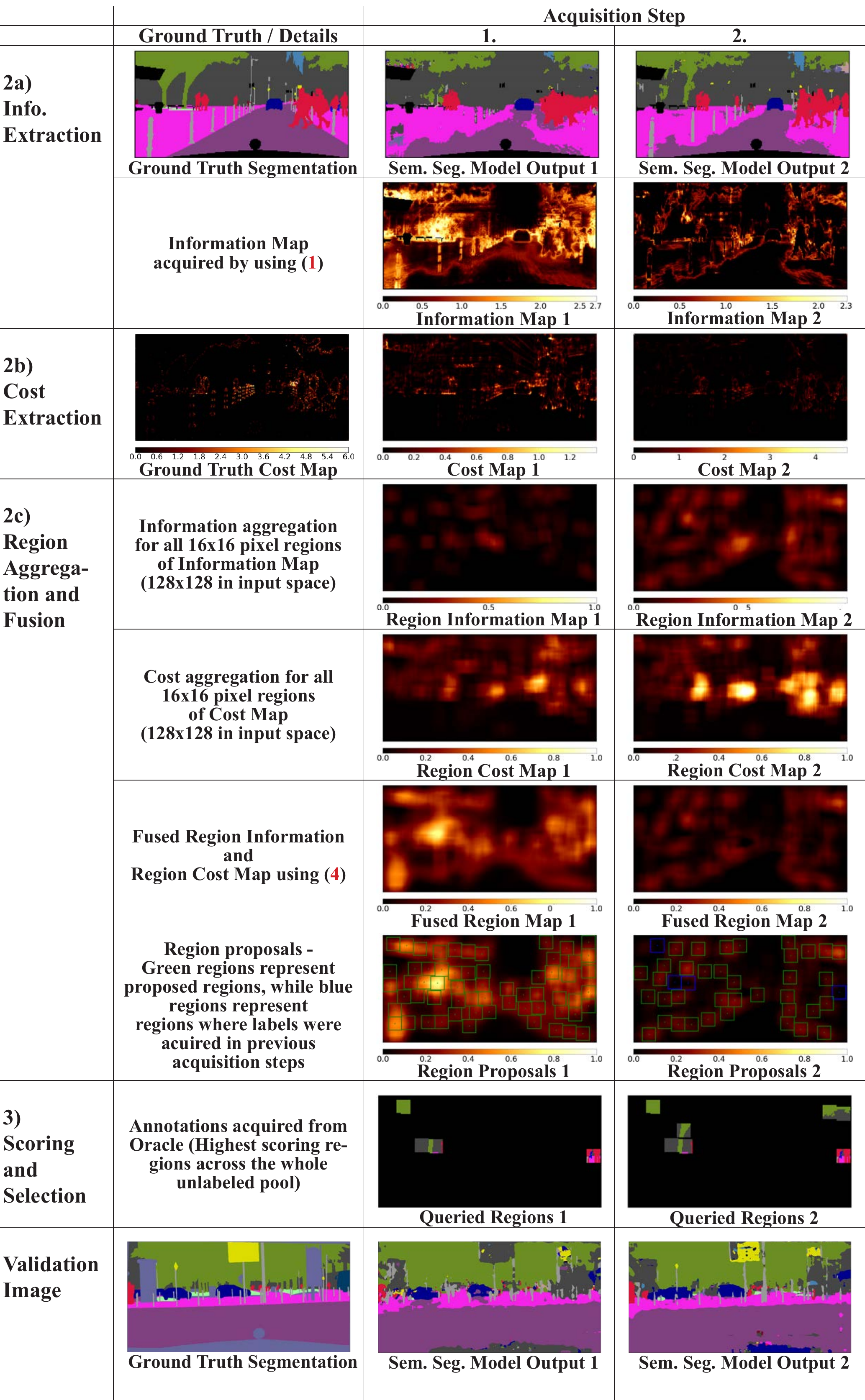}}
\end{subfigure}
\caption{Detailed overview of \textit{CEREALS}. First two acquisition steps. }
\end{figure}

\begin{figure}[htb!]
\centering
\begin{subfigure}{\includegraphics[width=0.91\textwidth]{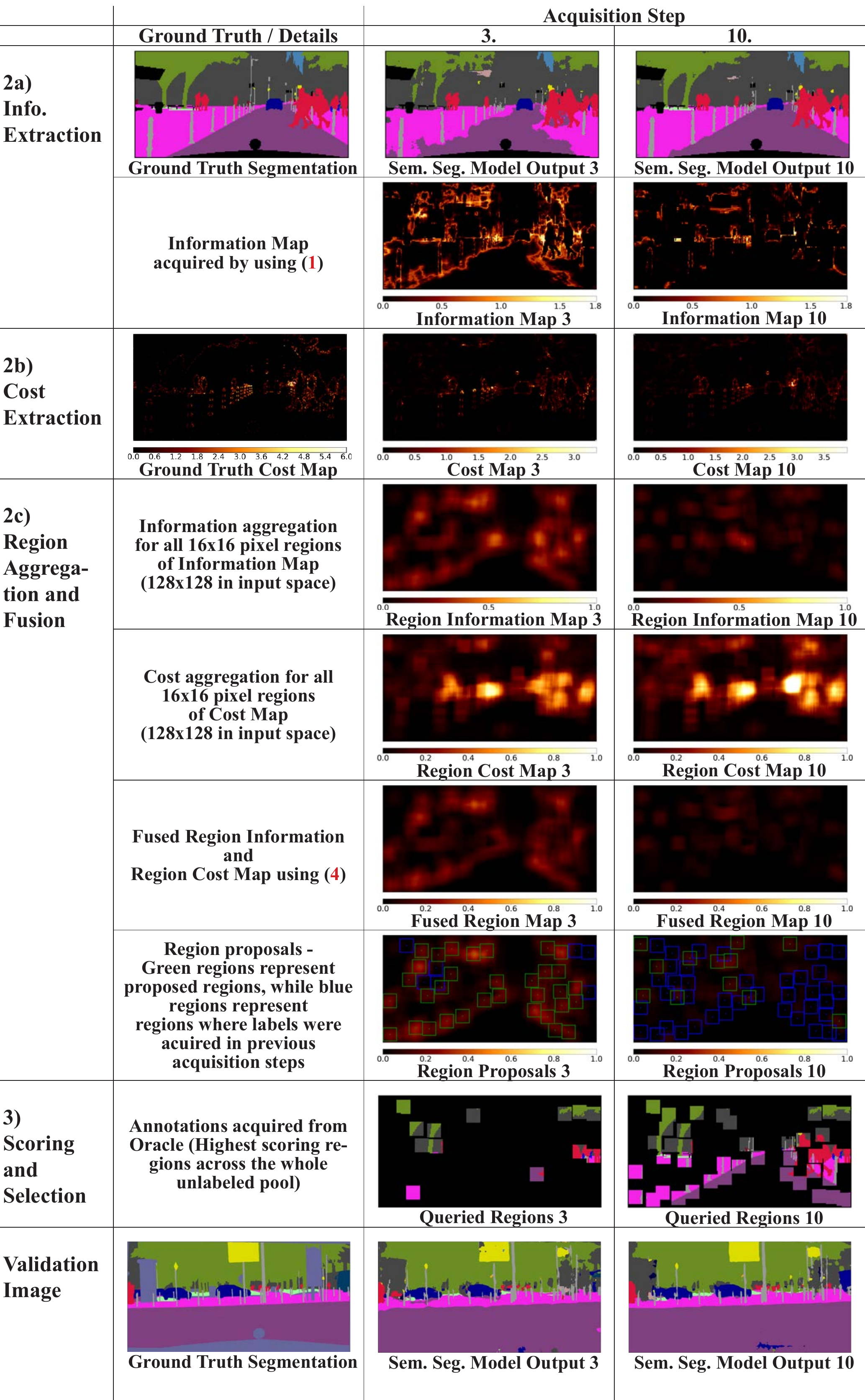}}
\end{subfigure}
\caption{Detailed overview of \textit{CEREALS}. Acquisition steps three and ten. }
\end{figure}

\end{document}